\documentclass[lettersize,journal]{IEEEtran}
\usepackage{amsmath,amsfonts}
\usepackage{algorithmic}
\usepackage{array}
\usepackage[caption=false,font=normalsize,labelfont=sf,textfont=sf]{subfig}
\usepackage{textcomp}
\usepackage{stfloats}
\usepackage{url}
\usepackage{verbatim}
\usepackage{graphicx}
\usepackage{amsmath}
\usepackage{amsthm}
\usepackage{mathrsfs}
\usepackage{setspace}
\usepackage{xcolor} 
\usepackage{amssymb}
\usepackage{pifont}
\usepackage[utf8]{inputenc}
\newcommand{\cmark}{\ding{51}}%
\newcommand{\xmark}{\ding{55}}%

\def\BibTeX{{\rm B\kern-.05em{\sc i\kern-.025em b}\kern-.08em
    T\kern-.1667em\lower.7ex\hbox{E}\kern-.125emX}}
\usepackage{balance}
\begin{document}
\title{How important are activation functions in regression and classification? A survey, performance comparison, and future directions}
\author{Ameya D. Jagtap$^*$ and George Em Karniadakis
\thanks{Division of Applied Mathematics, Brown University, 182 George Street, Providence, RI 02912, USA.  Corresponding Author $^*$ Emails: ameya$\_$jagtap@brown.edu, ameyadjagtap@gmail.com}}

\markboth{}
{How important are activation functions in regression and classification? A survey, performance comparison, and future directions}

\maketitle

\begin{abstract}
Inspired by biological neurons, the activation functions play an essential part in the learning process of any artificial neural network commonly used in many real-world problems. Various activation functions have been proposed in the literature for classification as well as regression tasks. In this work, we survey the activation functions that have been employed in the past as well as the current state-of-the-art. In particular, we present various developments in activation functions over the years and the advantages as well as disadvantages or limitations of these activation functions. We also discuss classical (fixed) activation functions, including rectifier units, and adaptive activation functions. In addition to discussing the taxonomy of activation functions based on characterization, a taxonomy of activation functions based on applications is presented. To this end, the systematic comparison of various fixed and adaptive activation functions is performed for classification data sets such as the MNIST, CIFAR-10, and CIFAR-100. In recent years, a physics-informed machine learning framework has emerged for solving problems related to scientific computations. For this purpose, we also discuss various requirements for activation functions that have been used in the physics-informed machine learning framework. Furthermore, various comparisons are made among different fixed and adaptive activation functions using various machine learning libraries such as TensorFlow, Pytorch, and JAX. Our findings show that activation functions such as ReLU and its variants, which are currently the state-of-the-art for many classification problems, do not work well in physics-informed machine learning frameworks due to the stringent requirement of the existence of derivatives, whereas other activation functions such as hyperbolic tangent, swish and sine give better performance, with superior results achieved with adaptive activation functions, especially for multiscale problems.
\end{abstract}

\begin{IEEEkeywords}
Activation function survey, Machine learning, Deep learning, Physics-informed machine learning, Neural networks
\end{IEEEkeywords}

\section{Introduction}\subsection{Aim}
In recent years, artificial neural networks (ANN) have achieved remarkable success in both academia as well as industry. The ANN has found success in a variety of industries, such as cybersecurity \cite{shaukat2020performance, sarker2021deep}, manufacturing technologies \cite{taha2011fuzzy, casalino2016ann}, healthcare \cite{wiens2018machine}, financial services \cite{coakley2000artificial}, food industry \cite{guine2019use}, and energy \cite{ahmad2014review}, to name a few, where they perform a range of tasks, including logistics, inventory management, etc. With various scientific applications, ANN is now the most sophisticated algorithm for understanding complicated sensory input, such as images \cite{turkmen2016ann}, video \cite{jiang2008no}, audio \cite{heckmann2002dct}, etc.  The first computational model for neural networks was created by McCulloch and Pitts \cite{mcculloch1943logical}, whereas the first ANN, also called as \textit{Perceptron} was invented by Rosenblatt \cite{rosenblatt1958perceptron}. Ivakhnenko and Lapa \cite{ivakhnenko1967cybernetics} released the \textit{Group Method of Data Handling}, which was the first functional neural network with multiple layers. However, a seminal book written by Minsky and Papert \cite{minsky1969introduction} showed that these early neural networks were incapable of processing simple tasks, and computers lacked the capability to process usable neural networks. Early neural networks developed by Rosenblatt only comprised one or two trainable layers. Such simplistic networks cannot mathematically represent complicated real-world phenomena. Deep neural networks (DNN) are ANNs that have a large number of trainable hidden-layers. The DNN was theoretically more versatile. However, without success, researchers spent several years trying to figure out how to train the DNN, as it could not scale with the first neural network's straightforward \textit{hill-climbing algorithm}. In a groundbreaking study, Rumelhart et al. \cite{rumelhart1985learning} created the \textit{backpropagation} training algorithm to train the DNN. The backpropagation algorithm is used to effectively train any ANN using a gradient descent method that takes advantage of the chain rule. Since then, ANN research has opened up all of the exciting and transformative developments in computer vision \cite{sebe2005machine}, speech recognition \cite{deng2013machine}, natural language processing \cite{chowdhary2020natural}, and even in scientific computations called Physics-Informed Machine Learning \cite{karniadakis2021physics}.

Artificial neurons are a set of connected units or nodes in an ANN that loosely replicate the neurons in a biological brain. The \textit{Threshold Logic Unit}, also known as the \textit{Linear Threshold Unit}, was the first artificial neuron and was first put forth by McCulloch and Pitts \cite{mcculloch1943logical} in 1943.
The model was intended to serve as a computational representation of the brain's '\textit{nerve net}'. Each artificial neuron receives inputs and generates a single output that can be delivered to a number of other neurons. Artificial neurons do not simply output the raw data they receive. Instead, there is one more step, which is analogous to the rate of action potential firing in the brain and is called an activation function. The introduction of the  activation function in ANN was inspired by biological neural networks whose purpose is to decide whether a particular neuron fires or not.  
The simple addition of such a nonlinear function can tremendously help the network to exploit more, thereby learning faster. There are various activation functions proposed in the literature, and it is difficult to find the optimal activation function that can tackle any problem.
In this survey, our aim is to discuss various advantages as well as disadvantages or limitations of classical (fixed) as well as modern (adaptive) activation functions.

\subsection{Contributions of this work}
 To the best of our knowledge, this is the first comprehensive survey of activation functions for both classification and regression problems. Apart from that, we also present several original contributions, summarized below.
\begin{itemize}
 \item First, we have done a comprehensive survey of the classical (fixed) activation functions like the real-valued activation functions, including the rectifier units. This also includes the oscillatory as well as the non-standard activation functions. We further discuss various properties of the classical activation functions that make them best suited for particular tasks.
 \item We present an applications-based taxonomy of activation functions, which is more suitable from a scientific computation point-of-view. Apart from real-valued activation functions, we also discuss various complex-valued activation functions, which have many applications in remote sensing, acoustics, opto-electronics, image processing, quantum neural devices, robotics, bioinformatics, etc. Furthermore, the quantized activations, which are a special type of activation function, are also discussed in detail. The quantized activations are typically used to improve the efficiency of the network without degrading the performance of the model.
 \item The state-of-the-art adaptive activation functions, which outperform their classical counterparts, are thoroughly discussed. Such adaptive activation not only accelerates the training of the network but also increases the prediction accuracy. Various adaptive activation functions such as stochastic or probabilistic, ensemble, fractional, etc. are discussed in detail. To this end, we also compare various fixed and adaptive activation functions systematically for classification data sets including MNIST, CIFAR-10, and CIFAR-100.
 \item Physics-informed machine learning is an intriguing approach that seamlessly incorporates the governing physical laws into the machine learning framework. Such incorporation of physical laws sets additional requirements for the activation function. To this end, we discuss these requirements for specifically solving scientific problems using a physics-informed machine learning framework. We used various fixed and adaptive activation functions to solve different PDEs. Furthermore, we also compare the predictive accuracy of the solution for different activation functions using various machine learning libraries such as TensorFlow, PyTorch, and JAX using clean and noisy data sets. A run-time comparison is made among the TensorFlow, PyTorch, and JAX machine learning libraries.
\end{itemize}

\subsection{Organization}
This paper is organized as follows: In Section 2, we provide a historical perspective of the activation functions. In section 3, we compare biological and artificial neurons in detail, followed by section 4, where some of the desired features of the neurons are discussed in depth. Section 5 gives a detailed discussion about the taxonomy of the activation functions. Section 6 covers several classical activation functions with their improved versions in detail. Section 7 is devoted to the motivation and historical development of complex-valued activation functions. Similarly, section 8 discusses the efficient quantized activations for quantized neural networks. In Section 9, we discuss various types of adaptive or trainable activation functions, ranging from stochastic or probabilistic to ensemble activations. Various fixed and adaptive activation functions are compared in terms of accuracy for the MNIST, CIFAR-10, and CIFAR-100 data sets in Section 10 using two convolutional neural network-based models, namely, \textit{MobileNet} and \textit{VGG16}. Section 11 presents a discussion on the activation function required for solving regression problems using a physics-informed machine learning framework. Finally, we summarize our findings in section 12.

\section{Activation functions: A historical perspective}
The activation function is a function that acts on the output of the hidden layer and the output layer (optional), which helps the neural network to learn complex features of the data. The motivation to employ the activation function came from the biological neural networks where this activation function decides whether the particular neuron activates (fires) or not. The important feature of the activation function is the ability to introduce nonlinearity in the network in order to capture nonlinear features, without which the neural network acts as a linear regression model. Cybenko \cite{cybenko1989approximation} and Hornik \cite{hornik1989multilayer} argue for the activation function's nonlinearity, demonstrating that the activation function must be bounded, non-constant, monotonically rising, and continuous to ensure the neural network's universal approximation property.

\begin{figure*}[tb]
\centering
\includegraphics[scale=0.95]{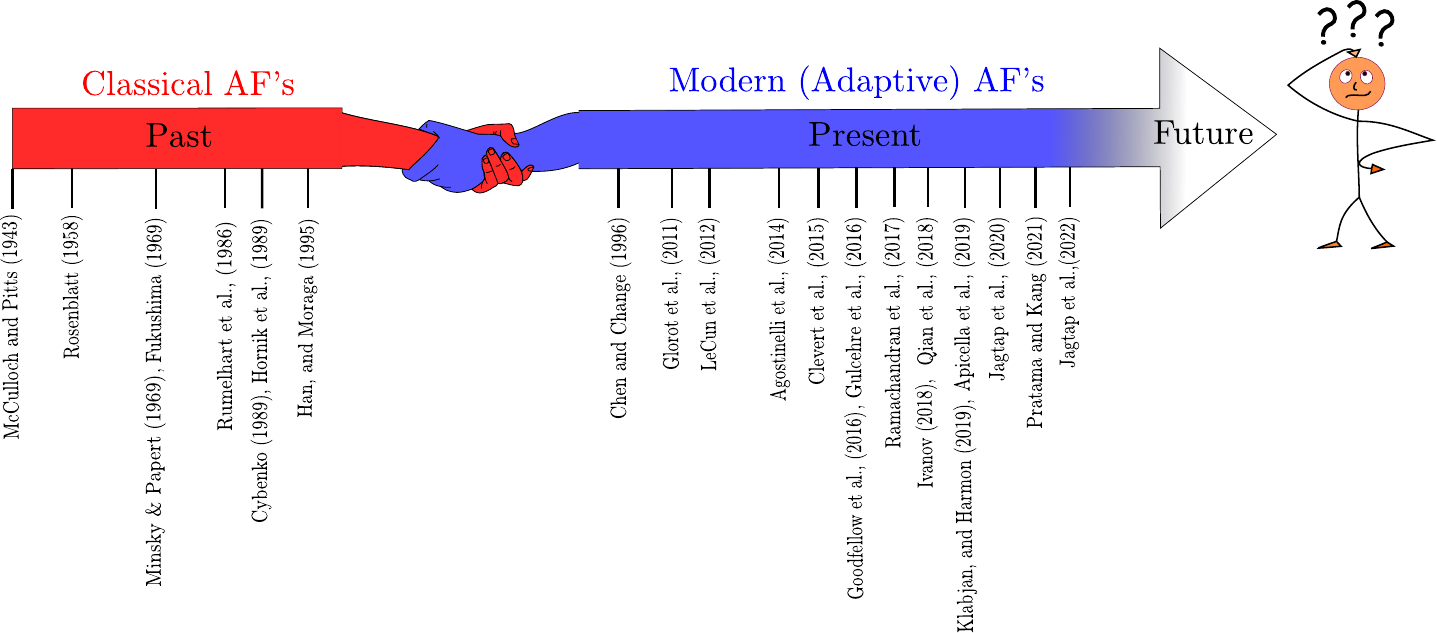}
\caption{Notable historical developments over the years related to the activation function.}
\label{fig:HisAF}
\end{figure*}

Figure \ref{fig:HisAF} shows some of the notable historical developments related to activation functions that we shall discuss here. In the literature, the term activation function has been referred to with different names, such as squashing function \cite{haykin1994neural}, and output function or transfer function \cite{duch1999survey}. In \cite{dasgupta1993power}, Dasgupta and Schnitger defined the activation as a real-valued functions defined on subset of $\mathbb{R}^n$. In \cite{goodfellow2016deep}, Goodfellow et al. described the activation function as a \textit{fixed nonlinear function}. The activation function was first used in the work of Fukushima \cite{fukushima1969visual} for visual feature extraction in hierarchical neural networks. Later, Fukushima and Miyake \cite{fukushima1982neocognitron} proposed it again for visual pattern recognition.
Hahnloser et al. \cite{hahnloser2000digital, hahnloser2003permitted} argued that the activation function is strongly related to biological neurons. The work of Glorot et al. \cite{glorot2011deep} describes the utility of the ReLU activation function. During the early 1990's, the Sigmoid activation function \cite{han1995influence} was one of the most popular activation functions. Due to its vanishing gradient problem, the notable improvement of Sigmoid such as the \textit{improved logistic Sigmoid function} \cite{qin2018optimized} has recently been proposed. In the late 1990's, researchers widely used the hyperbolic tangent function \cite{lecun2012efficient} as an activation function. Both Sigmoid and hyperbolic tangent face vanishing gradient problems and have difficulties with large inputs as they saturate for large input values. Later researchers proposed some alternative functions, and one of the most popular gradient vanishing-proof activation functions was the rectified linear unit (ReLU). As of 2017, the ReLU was the most popular activation function \cite{ramachandran2017searching} compared to widely used activation functions such as sigmoid and hyperbolic tangent functions. Although the ReLU was very successful in many applications such as speech recognition \cite{maas2013rectifier} and computer vision \cite{glorot2011deep}, it suffered from issues such as non-differentiability at zero, unboundedness, and the most famous, the \textit{dying ReLU problem} \cite{maas2013rectifier}. Several other linear as well as non-linear variants of the rectifier unit were proposed. The notable ones are leaky ReLU \cite{maas2013rectifier} and parametric ReLU \cite{he2015delving}, which are linear variants. Some of the non-linear variants are softplus \cite{dugas2000incorporating}, Exponential linear units (ELU) \cite{clevert2015fast}, Gaussian Error Linear Units (GELU) \cite{hendrycks2016gaussian}, Swish \cite{ramachandran2017searching} and Mish \cite{misra2019mish}.

Recent years have seen an increase in studies on adaptive activation functions. In the early studies, the generalized hyperbolic tangent function was suggested by Chen and Chang \cite{chen1996feedforward}. It is parameterized by two additional positive scalar values. Vecci et al. \cite{vecci1998learning} proposed an adaptive spline activation function. Trentin \cite{trentin2001networks} gives empirical evidence that learning the amplitude for each neuron is superior to having a unit amplitude for all activation functions  (either in terms of generalization error or speed of convergence). Goh et al. \cite{goh2003recurrent}  proposed a trainable amplitude activation function. The activation function adaptation algorithm is suggested for sigmoidal feed-forward neural network training by Chandra and Singh \cite{chandra2004activation}. Recently,  Agostinelli et al.  \cite{agostinelli2014learning} proposed learning activation functions. Eisenach et al. \cite{eisenach2016nonparametrically}  proposed a nonparametric method for learning activation functions. The algebraic activation functions were proposed by Babu and Edla \cite{naresh2017new}. Urban et al. proposed stochastic activation functions based on Gaussian processes in \cite{urban2017gaussian}. Alcaide et al. \cite{alcaide2018swish} proposed the E-swish activation function. The trained activation function was proposed by Ertu{\u{g}}rul \cite{ertuugrul2018novel}. Convolutional neural networks with adaptive activations were proposed by Qian et al. \cite{qian2018adaptive}. In order to discover the ideal quantization scale, Choi et al. \cite{choi2018pact} introduced PArameterized Clipping acTivation (PACT), which makes use of an activation clipping parameter that is tuned during training. Apicella et al. \cite{apicella2019simple} proposed an effective design for trainable activation functions. Several variants of adaptive ReLU activation function have been proposed recently, such as Parametric ReLU \cite{he2015delving}, S-shaped ReLU \cite{jin2016deep}, Flexible ReLU \cite{qiu2018frelu}, Paired ReLU \cite{tang2018joint}, etc. Similarly, adaptive ELUs like Parametric ELU \cite{shah2016deep}, Continuously Differentiable ELU \cite{barron2017continuously}, Shifted ELU \cite{grelsson2018improved}, Fast ELU \cite{qiumei2019improved}, Elastic ELU \cite{kim2020elastic}, etc. are also proposed in the literature. Recently, Jagtap et al. proposed a series of papers on global \cite{jagtap2020adaptive} and local \cite{jagtap2020locally} adaptive activation functions. The other way to introduce adaptivity is through the use of stochastic activation functions. Gulcehre et al. \cite{gulcehre2016noisy} proposed a noisy activation function where a structured bounded noise-like effect is added to allow the optimizer to exploit more and learn faster. On a similar idea, Shridhar \cite{shridhar2019probact} proposed the probabilistic activation functions, which are not only trainable but also stochastic in nature. In this work, the authors replicated uncertain behavior in the information flow to the neurons by injecting stochastic sampling from a normal distribution into the activations. The fractional activation functions with trainable fractional parameters are proposed by Ivanov \cite{ivanov2018fractional}. Later, Esquivel et al. \cite{zamora2019adaptive} provided adaptable activation functions based on fractional calculus. The ensemble technique is another way to adapt activation functions. The earlier work of Xu and Zhang \cite{xu2000justification} proposes an adaptive ensemble activation function consisting of sigmoid, radial basis, and sinusoid functions. In \cite{ismail2013predictions}, Ismail et al. select the most suitable activation function as a discrete optimization problem that involves generating various combinations of functions. Chen \cite{chen2016combinatorially} used multiple activation functions for each neuron for the problems related to stochastic control. Agostinelli et al. \cite{agostinelli2014learning} constructed the activation functions during the network training. Jin et al. \cite{jin2016deep} proposed the combination of a set of linear functions with open parameters. In recent years, a generalized framework, namely Kronecker Neural Networks for any adaptive activation functions, has been proposed by Jagtap et al. \cite{jagtap2022deep}. Good reviews of classical activation functions \cite{nwankpa2018activation, szandala2021review} and the modern trainable activation \cite{apicella2021survey} are available for various classification problems.

\section{Biological vs artificial neurons}
\begin{figure}[h!]
\centering
\includegraphics[ scale=0.5]{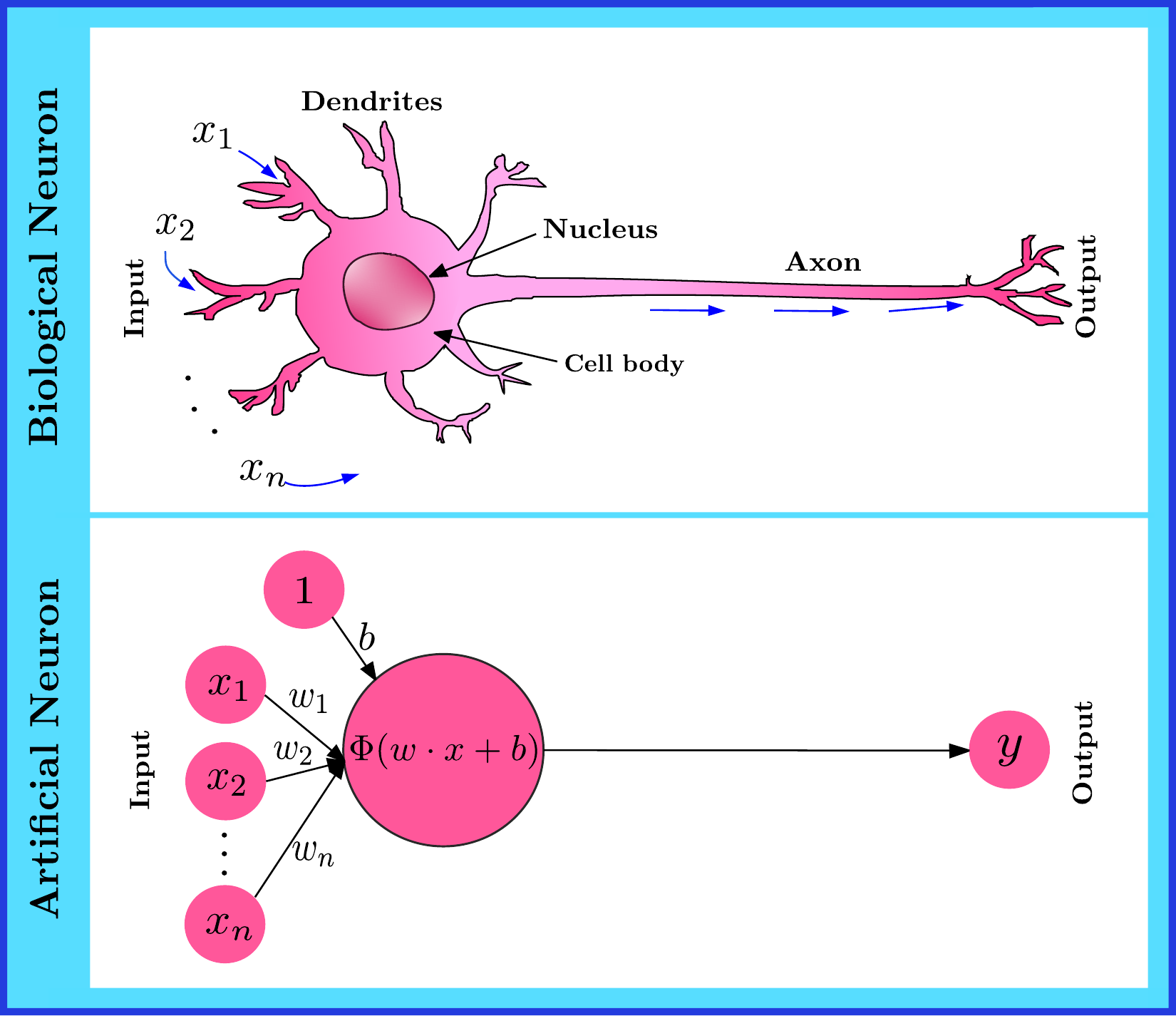}
\caption{Biological (top) vs artificial (bottom) neurons.}
\label{fig:BioArtNN}
\end{figure}

A neuron in the human brain is a biological cell that processes information. The biological neuron is depicted in the top figure \ref{fig:BioArtNN}. There are three basic elements of a biological neuron. The dendrites receive singles $x_i, i= 1,2,\cdots, n$ as an input from other neurons. A cell body (nucleus) that controls the activity of neurons and an axon that transmits signals to the neighboring neurons. The axon's length may be several times or even tens of thousands of times longer than the cell body. The axon is divided into various branches near its extremity, which are connected to the dendrites of other neurons. There are millions of massively connected neurons (around $10^{11}$), which is approximately equal to the number of stars in the Milky way \cite{brunak1990neural}. Every neuron is connected to thousands of neighboring neurons, and these neurons are organized into successive layers in the cerebral cortex.
Such a massively parallel neural network communicates through a very short train of pulses, in milliseconds, and has an ability that includes parallel processing, fast learning ability, adaptivity, generalization ability, very low energy consumption, and fault tolerance.  The artificial neuron in the bottom figure \ref{fig:BioArtNN} attempts to resemble the biological neuron. The artificial neuron, like a biological neuron, takes input $x_i, i= 1,2,\cdots, n$, and is made up of three basic parts: weights $w$ and bias $b$, which act as a dendrites, the activation function (denoted by $\Phi(\cdot)$) that act as a cell body and nucleus, and the output $y$. The artificial neuron has been generalized in many ways. Among all the components, the most obvious is the activation function. The activation function plays an important role in replicating the biological neuron firing behaviour above some threshold value. Unlike biological neurons, which send binary values, artificial neurons send continuous values, and depending on the activation function, the firing behavior of artificial neurons changes significantly.

\section{Desired characteristics of the activation functions}
There is no universal rule for determining the best activation function; it varies depending on the problem under consideration.
Nonetheless, some of the desirable qualities of activation functions are well known in the literature. The following are the essential characteristics of any activation function. 
\begin{enumerate}
\item \textbf{Nonlinerity}:  One of the most essential characteristics of an activation function is nonlinearity.
In comparison to linear activation functions, the non-linearity of the activation function significantly improves the learning capability of neural networks.
In \cite{cybenko1989approximation} Cybenko and \cite{hornik1989multilayer} Hornik advocates for the nonlinear property of the activation function, demonstrating that the activation function must be bounded, non-constant, monotonically growing, and continuous in order to ensure the neural network's universal approximation property. In \cite{morita1993associative, morita1996memory} Morita later discovered that neural networks with non-monotonic activation functions perform better in terms of memory capacity and retrieval ability. Recently, Sonoda and Murata \cite{sonoda2017neural} showed that neural networks equipped with unbounded but non-polynomial activation functions (e.g., ReLU) are universal approximators.

 \item \textbf{Computationally cheap}: The activation function must be easy to evaluate in terms of computation.
This has the potential to greatly improve network efficiency. 
 \item \textbf{The vanishing and exploding gradient problems}: The vanishing and exploding gradient problems are the important problems of activation functions. The variation of the inputs and outputs of some activation functions, such as the logistic function (Sigmoid), is extremely large.
To put it another way, they reduce and transform a bigger input space into a smaller output space that falls between [0,1]. As a result, the back-propagation algorithm has almost no gradients to propagate backward in the network, and any residual gradients that do exist continue to dilute as the program goes down through the top layers. Due to this, the initial hidden-layers are left with no information about the gradients. For hyperbolic tangent and sigmoid activation functions, it has been observed that the saturation region for large input (both positive and negative) is a major reason behind the vanishing of gradient. One of the important remedies to this problem is the use of non-saturating activation functions. Other non-saturating functions, such as ReLU, leaky ReLU, and other variants of ReLU, have been proposed to solve this problem. 
 
 \item \textbf{Finite range/boundedness}: Gradient-based training approaches are more stable when the range of the activation function is finite, because pattern presentations significantly affect only limited weights.
 \item \textbf{Differentiability}: The most desirable quality for using gradient-based optimization approaches is continuously differentiable activation functions. This ensures that the back-propagation algorithm works properly. 
\end{enumerate}

\section{Taxonomy of activation functions}

\subsection{Characterization based taxonomy}
A taxomomy of activation functions was proposed in \cite{apicella2021survey}, which we also follow in the current work. Specifically, the activation functions can be broadly divided into linear and nonlinear functions. For all realistic problems, nonlinear activation functions are often employed. 
\begin{figure}[!h]
\centering
\includegraphics[ scale=0.52]{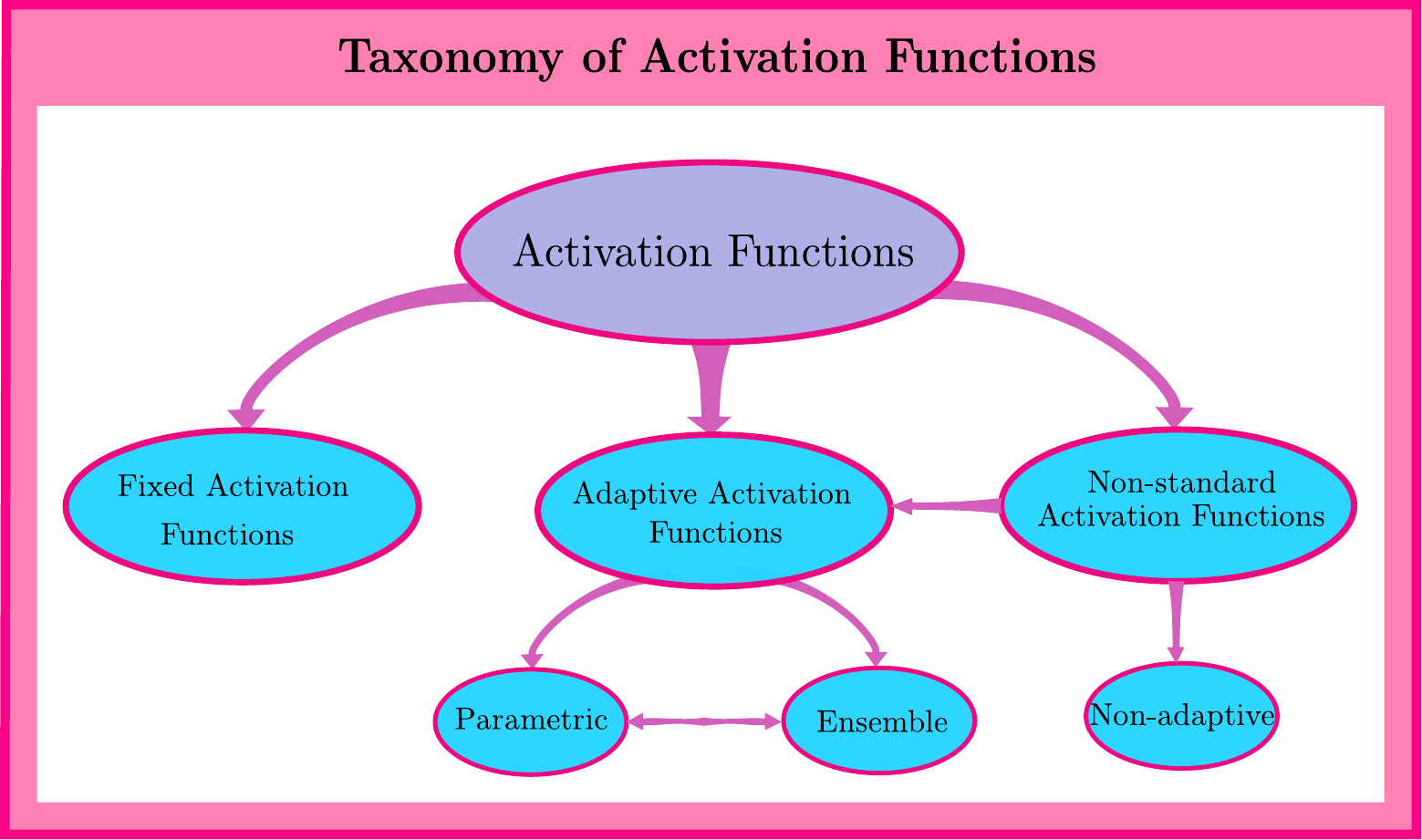}
\caption{Taxonomy based on characterization of activation functions: The activation functions are broadly divided into three categories, namely, fixed, adaptive, and non-standard activation functions. There are various ways to inject the adaptive nature of the activation function, like ensemble or parametric approaches. The non-standard activation functions can be fixed or adaptive.}
\label{fig:AF_Tax1}
\end{figure}
Figure \ref{fig:AF_Tax1} shows the taxonomy based on characterization of activation functions, which divides the activation function into three major categories, first is the fixed activation function (see Section \ref{Sec6}) that contains all the  fixed (classical) activations, including the rectifier units. The second category is the adaptive or modern activation functions that can adapt itself, which is further divided into parametric as well as ensemble activations. The third category is non-standard activation, such as multi-fold activation from previous layers, which may be fixed or adaptive.  

\subsection{Application based taxonomy}
\begin{figure}[!h]
\centering
\includegraphics[ scale=0.5]{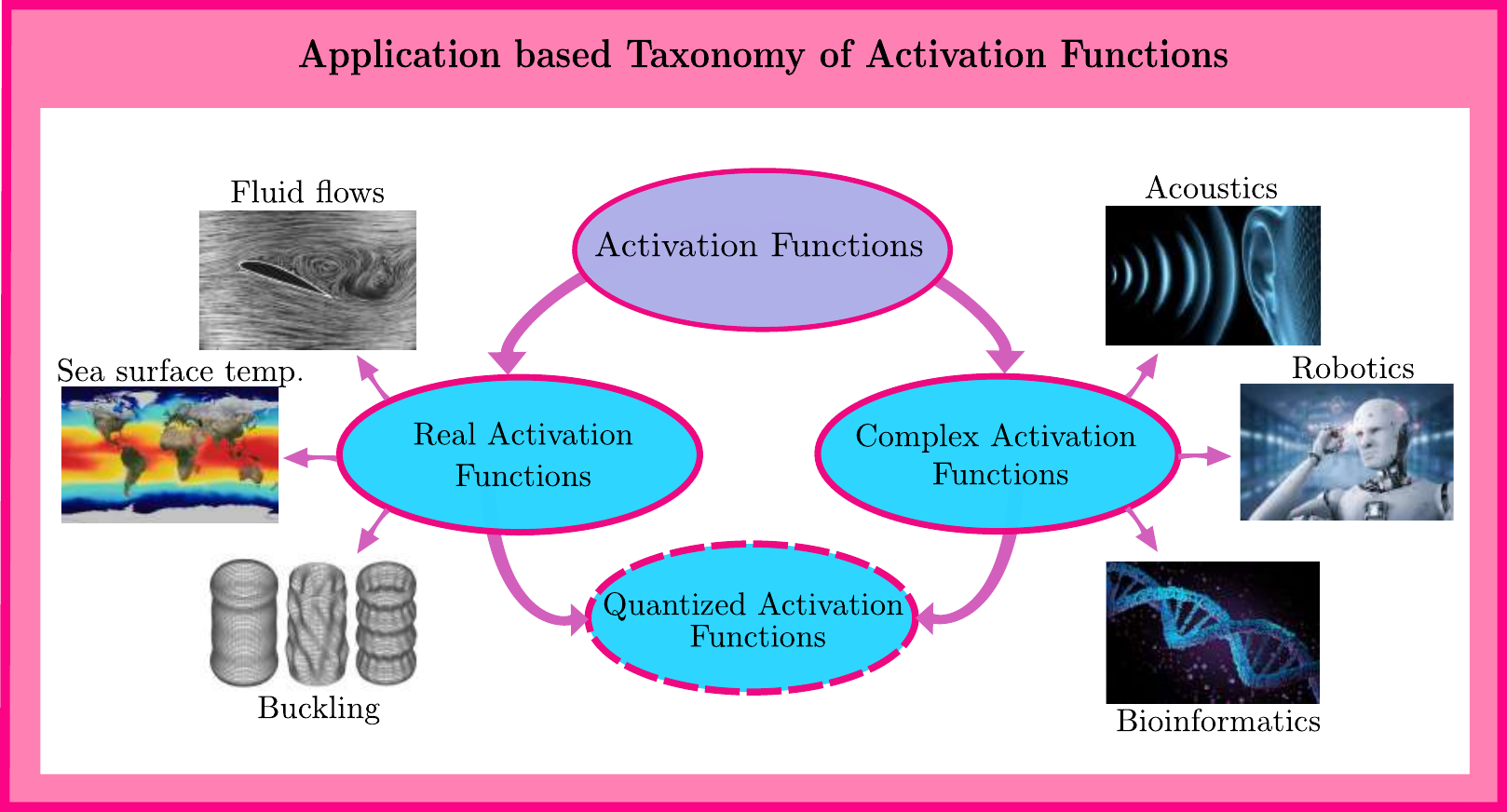}
\caption{Taxonomy based on the application of activation functions: The activation function can also be divided based on the application at hand and the nature of its output. In particular, we propose real and complex-valued activation functions. Few scientific applications of such activation functions are also shown in the taxonomy. For efficient computation, both real and complex-valued activation functions can be quantized.}
\label{fig:AF_Tax2}
\end{figure}

Although the taxonomy of activation functions based on their characterization encompasses all the activation functions, it is possible to define the taxonomy based on their applications to real-world problems; this can help us to further broaden our understanding of activation functions from a physical point-of-view. In particular, we are proposing an application-based taxonomy for scientific applications, where the output field states are real or complex-valued; for example, applications in acoustics, robotics, and bioinformatics may include complex arithmetic representation. A schematic representation of this is shown in Fig. \ref{fig:AF_Tax2}. The real-valued activations are well-known in the literature, and we shall discuss them in detail later. The complex-valued activation functions are another set of activation functions whose output is a complex number. These activation functions are often required due to the complex-valued output, which has many applications in science and engineering, such as bioinformatics, acoustics, robotics, opto-electronics, quantum neural devices, image processing, etc.
The process of mapping continuous values from an infinite set to discrete finite values is known as \textit{Quantization}. Both real and complex-valued activation functions can be quantized in order to reduce memory requirements. Quantized activations are the class of activation functions that is efficient in terms of memory requirements as well as increased compute efficiency. The output of the quantized activation is integers rather than floating point values.  Note that, just like real-valued activation functions, both complex-valued as well as quantized activations can be made adaptive by introducing the tunable variables. The upcoming sections provide a clear discussion of the various complex-valued and quantized activation functions. 

\section{Classical activation functions in artificial neural networks}\label{Sec6}
This section gives details about the various fixed activation functions that have been proposed in the literature. These activation functions do not contain any tunable parameters. The following are the most common fixed activation functions used.
\subsection{Linear and piece-wise linear functions} 
Linear function $ \Phi(x) = x $ is the simplest form of activation function. It has a constant gradient, and the descent is based on this constant value of gradient. The range of linear function is $(-\infty,\infty)$, and has a $C^1$ order of continuity.

The piece-wise linear activation can be defined as $$ \Phi(x) = \begin{cases}
 0 & x<-b, \\ x+b &  -b <x <b, \\ 1 &x>b,  \end{cases}$$ 
 where $b$ is a constant. The derivative of piece-wise linear activation is not defined at $x = \pm b$, and it is zero for $x <-b$ and $x>b$. The linear function has a $C^0$ order of continuity and a range of $[0,1]$. 

\subsection{Step function}  
It is also known as Heaviside or the unit step function, and is defined as
$$ \Phi(x) = \begin{cases}
        0 & \text{for}~x < 0 \\ 1 &\text{for}~ x \geq 0.
         \end{cases}
$$
Step function is one of the most basic forms of activation function. The derivative of a step function is zero when $x\neq 0$, and it is not defined when $x=0$. The step function has a $C^{-1}$ order of continuity and a $\{0,~1\}$ range. 

\subsection{Sigmoid function} 
The sigmoid function \cite{han1995influence}, also called the logistic function, was a very popular choice of activation function till the early 1990's. It is defined as
  $$\Phi(x) = \frac{1}{1 + e^{-x}}.$$
In \cite{hinton2012deep}, Hinton et al. used a sigmoid activation function for automatic speech recognition. The major advantage of sigmoid activation is its boundedness. The disadvantages are: the vanishing gradient problem, the output not being zero-centered, and the saturation for large input values. In \cite{nair2010rectified}, Nair and Hinton showed that as networks became deeper, training with sigmoid activations proved less effective.  The range of the sigmoid function is [0,1] and has a $C^{\infty}$ order of continuity.

Some improvements to Sigmoid functions were proposed in the literature as follows.
\begin{itemize}
\item The Bipolar-Sigmoid activation function \cite{kwan1992simple} is defined as $$\Phi(x) = \frac{1 - e^{-x}}{1 + e^{-x}},$$
which has a range of [-1,1]. Bipolar sigmoid activation function is widely used in \textit{Hopfield neural network}, see for more details in \cite{mansor2016activation}.
\item The Elliott function defined in \cite{elliott1993better} has a similar structure as the Sigmoid function and it is given by
$$ \Phi(x) = \frac{0.5x}{1+|x|} + 0.5,$$ with range [0,1]. See \cite{farzad2019comparative} for more details.
 \item The parametric sigmoid function \cite{chandra2004activation} is given by $$\Phi(x) = \left(\frac{1}{(1+\text{exp}(-x))}\right)^m,~~ m\in (0,\infty).$$
The derivatives for $m \neq 1$ are skewed and their maxima shift from the point corresponding to the input value equal to zero.
\item An improved logistic Sigmoid function is proposed by \cite{qin2018optimized} to overcome the vanishing gradient problem as
$$ \Phi(x) = \begin{cases}
              \alpha (x-a) + \text{Sigmoid}(a), & x\geq a, \\ \text{Sigmoid}(x), & -a< x< a, \\ \alpha (x+a) + \text{Sigmoid}(a), & x\leq a,
             \end{cases}
$$ which has a [$-\infty, \infty$] range.
\item The scaled sigmoid function \cite{eger2019time} is defined as $$ \Phi(x) = \frac{4}{1+\text{exp}(-x)} - 2,$$ which has a range from [-2,2].
\end{itemize}

\subsection{Hyperbolic tangent (tanh) function} 
The tanh activation function is defined as
 $$\Phi(x) = \frac{e^x - e^{-x}}{e^x + e^{-x}}.$$
From the late 1990's till early 2000s, tanh was extensively used to train neural networks, and was a preferred choice over the classical sigmoid activation function. 
 The tanh activation has a range of $[-1,~1]$, and in general, is mostly used for regression problems.
 It has an advantage due to the zero-centered structure.
 The main problem with the tanh activations is the saturation region. Once saturated, it is really challenging for the learning algorithm to adapt the parameters and learn faster. This problem is the vanishing gradient problem.
 
 Some improved tanh activation functions have been proposed to eliminate the problems related to the original tanh activation function.
 \begin{itemize}
  \item A scaled hyperbolic tangent function is defined by \cite{lecun1998gradient} as
$$ \Phi(x) = A ~\tanh(S~x).$$
Here $A$ is the amplitude of the function, and $S$ determines its slope at the origin. The output of the activation function is the range $[-A,A]$. 
\item A rectified hyperbolic secant activation function \cite{samatin2016novel} is proposed as $\Phi(x) = x~ \text{sech}(x)$.
\item The Hexpo function \cite{kong2017hexpo} has a similar structure to the hyperbolic tangent function, which is defined as
$$ \Phi(x) = \begin{cases}
              -a ~(\text{exp}(-x/b)-1), & x \geq 0, \\ c ~(\text{exp}(-x/d)-1), & x < 0,
             \end{cases}
$$ and has range from [$-c,a$].

\item The penalized hyperbolic tangent function \cite{eger2019time} is defined as $$ \Phi(x) = \begin{cases}
                                                                                                                    \tanh(x) , & x\geq 0, \\ a ~ \tanh(x) & x<0,
                                                                                                                   \end{cases}
$$ which gives output in the range [$-a, 1$]. 
\item The linearly scaled tanh (LiSHT) activation function  \cite{roy2019lisht} is defined as $\Phi(x) = x ~\tanh(x)$.
The LiSHT scales the hyperbolic tangent function linearly to tackle its gradient diminishing problem.
 \end{itemize}

\subsection{Rectified Linear Unit (ReLU)} 
ReLU was primarily used to overcome the vanishing gradient problem. ReLU is the most common activation function used for classification problems. It is defined as
  $$ \Phi(x) = \begin{cases}
               0 & \text{for}~x \leq 0 \\ x & \text{for}~ x > 0.
              \end{cases}
$$
The derivative of ReLU is zero when $x< 0$, unity  when $x>0$, and at $x=0$, the derivative is not defined. The ReLU function has a range from $[0,~\infty)$ and has a $C^{0}$ order of continuity.
Apart from overcoming the vanishing gradient problem, the implementation of ReLU is very easy and thus cheaper, unlike tanh and sigmoid, where an exponential function is needed.
 Despite having some advantages over classical activations, ReLU still has a saturation region, which can prevent the learning of the networks. In particular, ReLU always discards the negative values. This makes the neurons stop responding to the gradient-based optimizer. This problem is known as \textit{dead or dying ReLU problem} \cite{maas2013rectifier, lu2019dying}, meaning the neurons stop outputting other than zero. This is one of the serious problems for ReLU, where most of the neurons become dead, especially when using a high learning rate.
 To overcome these problems, various variants of ReLU have been proposed.
 \begin{itemize}
  \item Leaky ReLU \cite{maas2013rectifier}: The leaky ReLU is defined as
    $$ \Phi(x) = \begin{cases}
               \alpha x & \text{for}~x \leq 0 \\ x & \text{for}~ x > 0.
              \end{cases}
$$
The hyperparameter $\alpha$ defines the leakage in the function (slope of the function).
By adding the small slope in the region $x<0$, the leaky ReLU overcomes the dying ReLU problem. Moreover, it has all the advantages of ReLU activation. One of the disadvantage of leaky ReLU is the hyperparameter $\alpha$, which needs to be defined appropriately. In most of the cases $\alpha = 0.01$ is used. In \cite{xu2015empirical}, Xu et al. compared the ReLU and leaky ReLU activation with the conclusion that the latter outperforms the former always.

\item Randomized leaky ReLU \cite{xu2015empirical}: In this case, the $\alpha$ is picked randomly in a given range during training and is fixed to its average value during testing.
\item Mirror ReLU \cite{zhao2016suppressing}: It is defined as
$$\Phi(x) = \begin{cases}
             1+x, & -1\leq x \leq 0 \\ 1+x,  & 0< x \leq 1 \\ 0,  &\text{Otherwise},
            \end{cases}
$$ which can also be defind as $\Phi(x) = \text{min}(\text{ReLU}(1-x), \text{ReLU}(1+x))$. 

\item Concatenated ReLU \cite{shang2016understanding}: The concatenated ReLU (CReLU) is given as
$$\Phi(x) = [\text{ReLU}(x),\text{ReLU}(-x)],$$
which has range of [$0,\infty$). The disadvantage of CReLU over ReLU is increased model complexity.
\item Elastic ReLU \cite{jiang2018deep}: The Elastic ReLU is defined as
$$ \Phi(x) = \text{max}(Rx,0),$$ where $R$ is the random number.
\item Bounded ReLU \cite{liew2016bounded}: The ReLU function has unbounded outputs for non-negative inputs. The bounded version of the ReLU function is defined as

$$\Phi(x) = \text{min}(\text{max}(0,x), A),$$ where $A$ is the maximum output value the function can produce.
\item V-shaped ReLU \cite{hu2018vrelu}: It is defined as
$$\Phi(x) = \begin{cases}
             x, & x\geq 0 \\ -x, & x<0,
            \end{cases}
$$
\item Dual ReLU \cite{godin2018dual}: The Dual ReLU has two dimensions, and it is defined as
$$\Phi(x) = \text{max}(0,a)- \text{max}(0,b),$$
where $a,b$ are the inputs in two different dimensions. The range of Dual ReLU is $(-\infty,\infty)$.
\item Randomly translated ReLU \cite{cao2018randomly}: It is given by
$$\Phi(x) = \begin{cases}
             x + \alpha, & x+\alpha > 0 \\ 0, & x+\alpha \leq 0,
            \end{cases}
$$ which has range of [$0,\infty$), where $\alpha$ is randomly sampled from Gaussian distribution.
\item Displaced ReLU (DReLU) \cite{macedo2019enhancing}: DReLU is a diagonally displaced ReLU function that generalizes both ReLU and SReLU \cite{clevert2015fast} by allowing its inflection to move diagonally from the origin to any point of the form. It is defined as
$$\Phi(x) = \begin{cases}
             x, & x\geq -\delta \\ -\delta, & x <-\delta,
            \end{cases}
$$ with range [$-\delta, \infty$). If $\delta = 0$ DReLU becomes ReLU, and if $\delta = 1$ DReLU becomes SReLU. 
\item Natural-Logarithm ReLU \cite{liu2019natural}: The Natural-Logarithm ReLU uses a logarithmic function to modify ReLU output for positive input values that increase the degree of nonlinearity. It is defined as
$$\Phi(x) = \text{ln}(\beta~\text{max}(0,x)+1). $$ This activation has output range from 0 to $\infty$, and the $\beta$ is constant.

\item Average Biased ReLU \cite{dubey2021average}: The Average-Biased ReLU is given by
$$\Phi(x) = \begin{cases}
             x-\beta, & x-\beta \geq 0 \\ 0, &\text{Otherwise},
            \end{cases}
$$ which has range of [$0,\infty$), and the variable $\beta$ is the average of input activation map to activation function.

 \end{itemize}
\begin{figure*}
\centering
\includegraphics[trim=1cm 1.4cm 1cm 1cm,  scale=0.5]{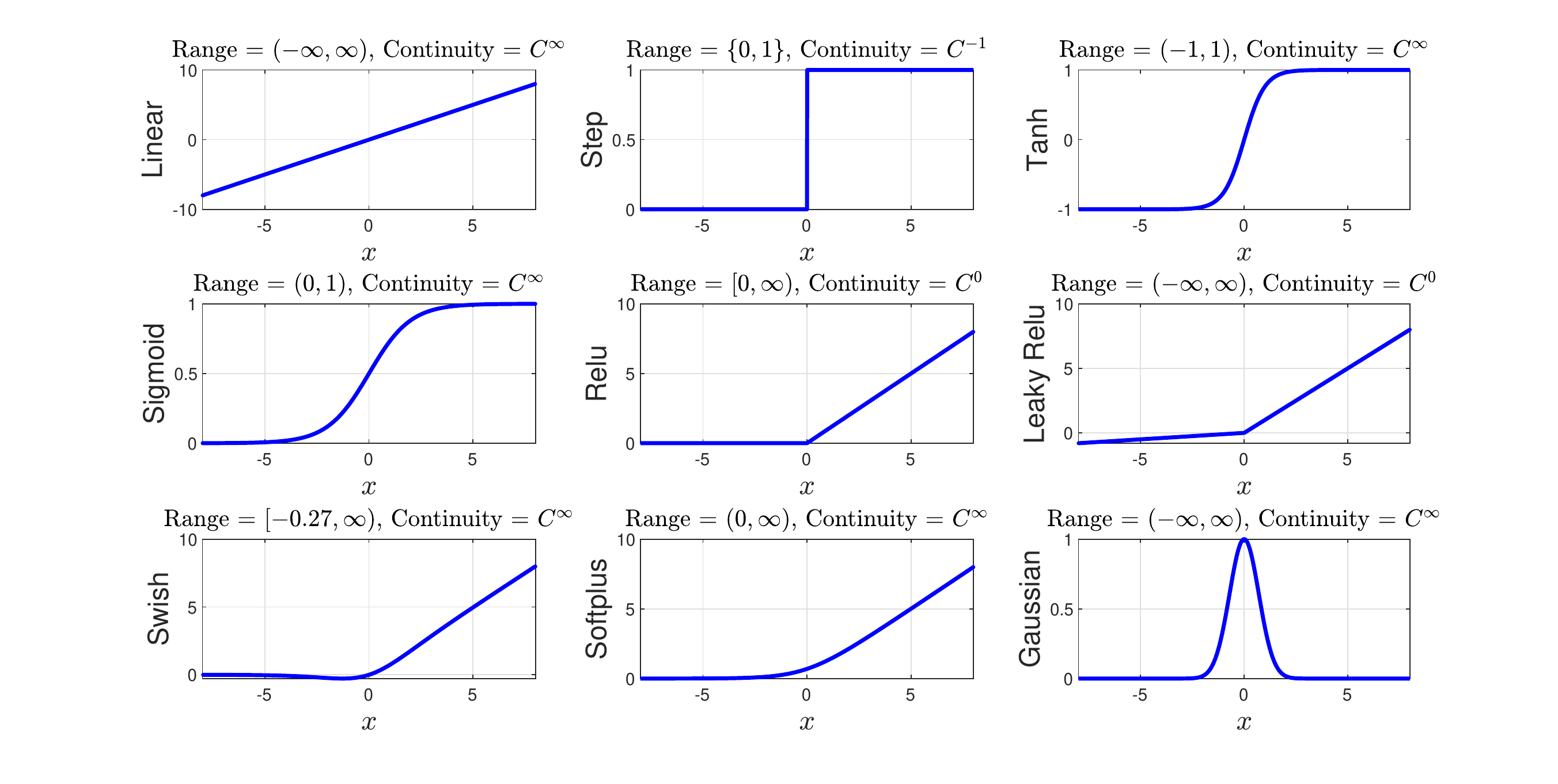}
\caption{Classical (fixed) activation functions with their range and continuity.}
\label{fig:AF_std}
\end{figure*}

\begin{figure*}
\centering
\includegraphics[trim=1cm 1.4cm 1cm 1cm,  scale=0.5]{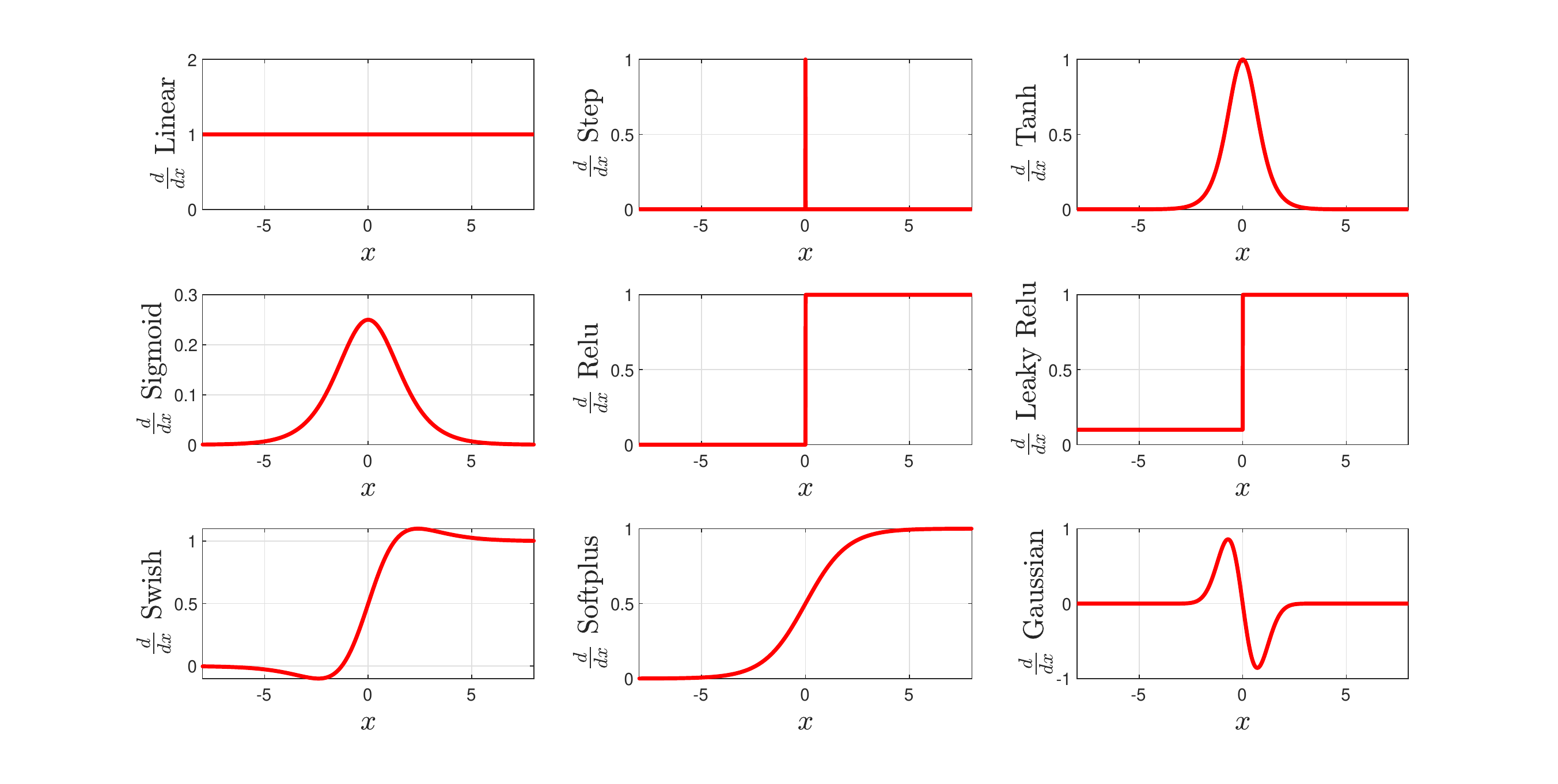}
\caption{Derivatives of classical activation functions.}
\label{fig:DAF_std}
\end{figure*}

\subsection{Gaussian Error Linear Units (GELUs)} 
In GELU activation \cite{hendrycks2016gaussian} the neuron input $x$ is multiplied by $m \sim \text{Bernoulli}(P(X \leq x))$ where $X \sim \mathcal{N}(0,1)$. The reason for choosing the Bernoulli distribution is that the neuron’s input follows a normal distribution, especially after batch normalization. The output of any activation function should be deterministic and not stochastic. So, the expected value of the transformation can be found as
$$\mathbb{E}(xm) = x \mathbb{E} = x P(X \leq x).$$
Since $P(X \leq x)$ is a cumulative distribution of Gaussian distribution, it is frequently computed with the error function.
Thus, GELU activation is defined as
$$\Phi(x) = x ~[1 + \text{erf}(x/ 2)],$$
which can be approximated as $0.5x(1 + tanh[ \sqrt{2/\pi}(x + 0.044715~x^3 )])$.

To improve its capacity for bidirectional convergence, the GELU was upgraded to the symmetrical GELU (SGELU) by Yu and Su \cite{yu2019symmetrical}. 
 
\subsection{Softplus function} The softplus function \cite{dugas2000incorporating} approximates the ReLU activation function in a smooth way, and it is defined as
 $$ \Phi(x) = \text{ln}(1+\text{exp}(x)),$$
 Softplus function is infinitely differentiable, and it has a range from $(0, \infty$).  
 In \cite{liu2016noisy}, Liu and Fuber proposed the noisy softplus activation. With Noisy Softplus, which is well-matched to the response function of LIF (Leaky Integrate-and-Fire) neurons, the performance of spiking neural networks can be improved, see \cite{liu2017noisy}. 
 The authors of \cite{liu2016noisy} proposed the following formula:
 $$ \Phi(x) = k \sigma~ \text{ln}\left(1+\text{exp}\left(\frac{x}{k\sigma}
 \right)\right),$$
 where $x$ is the mean input, $\sigma$ defines the noise level, and $k$ controls the curve scaling that can be determined by the neuron parameters.
 
\subsection{Exponential linear unit }
The Exponential linear unit (ELU) was first proposed in \cite{clevert2015fast} by Clevert et al., where they show that ELU outperforms all variants of ReLU with reduced training time as well as better accuracy in testing. The ELU is defined as
     $$ \Phi(x) = \begin{cases}
               \alpha (\text{exp}(x)-1) & \text{for}~x \leq 0 \\ x & \text{for}~ x > 0.
              \end{cases}
$$
When $x<0$, it takes on negative values, allowing the unit's average output to be closer to zero and alleviating the vanishing gradient problem. Also, due to the non-zero gradient for $x<0$, ELU does not suffer from the problem of dead neurons. 
Unlike ReLU, ELU bends smoothly at origin, which can be beneficial in the optimization process. Similar to leaky ReLU and parametric ReLU, ELU gives a negative output, which pushes the mean value towards zero. Again, $\alpha$ is a parameter which needs to be specified. For $\alpha = 1$, the function is smooth everywhere, which in turn helps the gradient descent algorithm to speed up. 

In \cite{klambauer2017self}, Klambauer et al. proposed the Scaled ELU (SELU) activation function, where the authors show that for the neural network consisting of a stack of dense layers, the network will \textit{self-normalize} if all the hidden-layers use the SELU activation function. However, there are some conditions for self-normalization; see \cite{klambauer2017self} for more details.

\begin{table*} 
\centering
\footnotesize
\begin{tabular}{|c|c|c|c|} 
\hline
\textbf{Activation Function} ($\mathbf{\Phi(x)}$) & \textbf{Derivatives} ($\mathbf{\Phi'(x)}$) & \textbf{Range} & \textbf{Continutity} \\ 
 \hline
Linear : $\Phi(x) = x$ & $\Phi'(x) =1 $ & ($-\infty,\infty$) & $C^{\infty}$ \\ \hline
Step : $\Phi(x) = \begin{cases}
                   0 & x\leq 0, \\ 1 & x>0
                  \end{cases} $ & $\Phi'(x) = \begin{cases}
                   0 & x\neq 0, \\ \text {Not defined}& x=0
                  \end{cases}$ & ($0,1$) & $C^{-1}$ \\
\hline
Sigmoid or Logistic : $\Phi(x) = \frac{1}{1+e^{-x}}$ & $\Phi'(x) =\Phi(x) (1-\Phi(x)) $ & ($0,1$) & $C^{\infty}$ \\ \hline
Rectifier Unit (ReLU) : $\Phi(x) = \begin{cases}
                   0 & x\leq 0, \\ x & x>0
                  \end{cases}$ & $\Phi'(x) =\begin{cases}
                   0 & x<  0, \\ 1 & x>0, \\  \text {Not defined}& x=0
                  \end{cases} $ & [$0,\infty$) & $C^{0}$ \\ \hline
Hyperbolic Tangent : $\Phi(x) = \frac{e^x-e^{-x}}{e^x+e^{-x}}$ & $\Phi'(x) = 1 - \Phi(x)^2 $ & ($-1,1$) & $C^{\infty}$ \\ \hline
Softplus : $\Phi(x) = \text{ln}(1+e^x)$ & $\Phi'(x) =~ \text{Sigmoid} $ & ($0,\infty$) & $C^{\infty}$ \\ \hline
Leaky Rectifier Unit (Leaky ReLU) : $\Phi(x) = \begin{cases}
                   0.01 x & x< 0, \\ x & x\geq 0
                  \end{cases}$ & $\Phi'(x) = \begin{cases}
                   0.01 & x<  0, \\ 1 & x>0, \\  \text {Not defined}& x=0
                  \end{cases} $ & ($-\infty,\infty$) & $C^{0}$  \\ \hline
Exponential Linear Unit (ELU) : $\Phi(x) = \begin{cases}
                   \alpha (e^x -1) & x\leq 0, \\ x & x>0
                  \end{cases}$ & $\Phi'(x) =  \begin{cases}
                   \alpha e^x & x<  0, \\ 1 & x>0, \\  \alpha & x=0
                  \end{cases} $  & ($-\alpha,\infty$) & $ \begin{cases}
 C^{1} & \text{If} \alpha = 1 \\ C^0 & \text{Otherwise} \end{cases}$\\ \hline
Gaussian : $\Phi(x) = e^{-x^2}$ & $\Phi'(x) = -2x \Phi(x)$ & ($0,1$] & $C^{\infty}$ \\ \hline
Swish ($\beta = 1$) : $\Phi(x) = x \cdot \text{Sigmoid}$ & $\Phi'(x) = x \cdot \Phi(x) (1-\Phi(x)) + \text{Sigmoid}$ & [$0,.\infty$) & $C^{\infty}$ \\ \hline
Oscillatory : $\Phi(x) = \sin(x)$ & $\Phi'(x) = -\cos(x)$ & [$-1,1$] & $C^{\infty}$ \\ \hline
\end{tabular}
\caption{Various classical activation functions, their derivatives, range, and order of continuity.}
\label{af_tABLE}
\end{table*}

\subsection{Mish function}
The \textit{Mish} \cite{misra2019mish} is a self-regularized, non-monotonic activation function defined as $$\Phi(x) = x \cdot  \tanh(\text{softplus}(x)).$$ While not evident at first sight, Mish is closely related to Swish \cite{ramachandran2017searching}, as its first derivative can be written in terms of the Swish function. Similar to Swish, Mish is unbounded above and bounded below. It is non-monotonic, has $C^{\infty}$ continuity, and is also self-gated.

\subsection{Radial activation functions}
 The traditional \textit{radial basis function (RBF)} neural networks uses the Gaussian function, which is given as
$$ \Phi(x) = e^{-x^2}.$$ The RBF neural networks with Gaussian activation functions have previously been successfully applied to a variety of difficult problems such as function approximation \cite{hartman1990layered, leonard1992using}, classification \cite{er2002face, savitha2012metacognitive} , etc. In \cite{zhang2016comprehensive}, Zhang and Suganthan found that radial basis function always leads to a better performance compared to hardlim and sign activation functions for UCI datasets using Random Vector Functional Link Networks (RVFL) \cite{pao1992neural}.
Other radial activation functions include \textit{Multiquadratics} \cite{lanouette1999process}, and \textit{Polyharmonic splines}.  The polyharmonic spline is a linear combination of polyharmonic radial basis functions and can be used as an activation function \cite{hryniowski2018polyneuron}.

Table \ref{af_tABLE} summarises the comparison of different activation functions, their derivatives, range, and order of continuity. The existence of the derivatives is an important feature for the backpropagation algorithm.
Figures \ref{fig:AF_std} and \ref{fig:DAF_std} show some of the classical activation functions and their derivatives, respectively.

\subsection{Wavelet activation functions}
Wavelet neural networks \cite{zhang1995wavelet} use wavelets as the activation function. Such networks have been utilized with some success in classification and identification problems. Moreover, these networks are effective because they can identify important features in high-frequency signals. A good review of wavelet networks and wavelets as activation functions can be found in \cite{thuillard2002review}. The Gaussian-wavelet type activation function is a class of piecewise linear, non-monotone activation functions \cite{he2015delving}, that have been employed in \cite{ji2019research, liu2022multistability}.

\subsection{Oscillatory activation functions}
 In \cite{gidon2020dendritic}, Gidon et al. found a new form of neuron in the human cortex that can learn the XOR function individually (a task that is difficult with single neurons employing other standard activation functions such as sigmoidal, ReLU, leaky ReLU, Softplus, ELU, Swish, Mish activations, etc). This is due to the fact that the zeros of the activation function for which $\Phi(x) = 0$ is the decision boundary for a neuron that emits an activation  $\Phi(x) = \Phi(w\cdot x + b)$. If the activation function $\Phi(x)$ has only one zero, then the decision boundary is a single hyperplane $x = w\cdot x + b = 0$. Two hyperplanes are required to distinguish the classes in the XOR data set \cite{minsky1969introduction}, hence activation functions with multiple zeros, as described in \cite{lotfi2014novel}, must be considered. The oscillatory activation functions fits this criterion perfectly. It is interesting to note that oscillatory activation functions were already present in the literature before the paper by Gidon et al. \cite{gidon2020dendritic}. In \cite{nakagawa1999chaos}, Nakagawa proposed the chaos neural network model applied to the chaotic autoassociation memory using sinusoidal activation function. Mingo et al. \cite{mingo2004fourier} proposed the Fourier neural network with sinusoidal activation function, see also Gashler and Ashmore \cite{gashler2014training}. In \cite{parascandolo2016taming}, Parascandolo et al. used sine activation for deep neural networks. For learning the XOR function with a single neuron, Noel et al. \cite{noel2021growing} proposed the Growing Cosine Unit (GCU) $\Phi(x) = x\cdot \cos(x)$.
 \begin{figure}
\centering
\includegraphics[trim=0.8cm 1.4cm 0.8cm 0.8cm, scale=0.43]{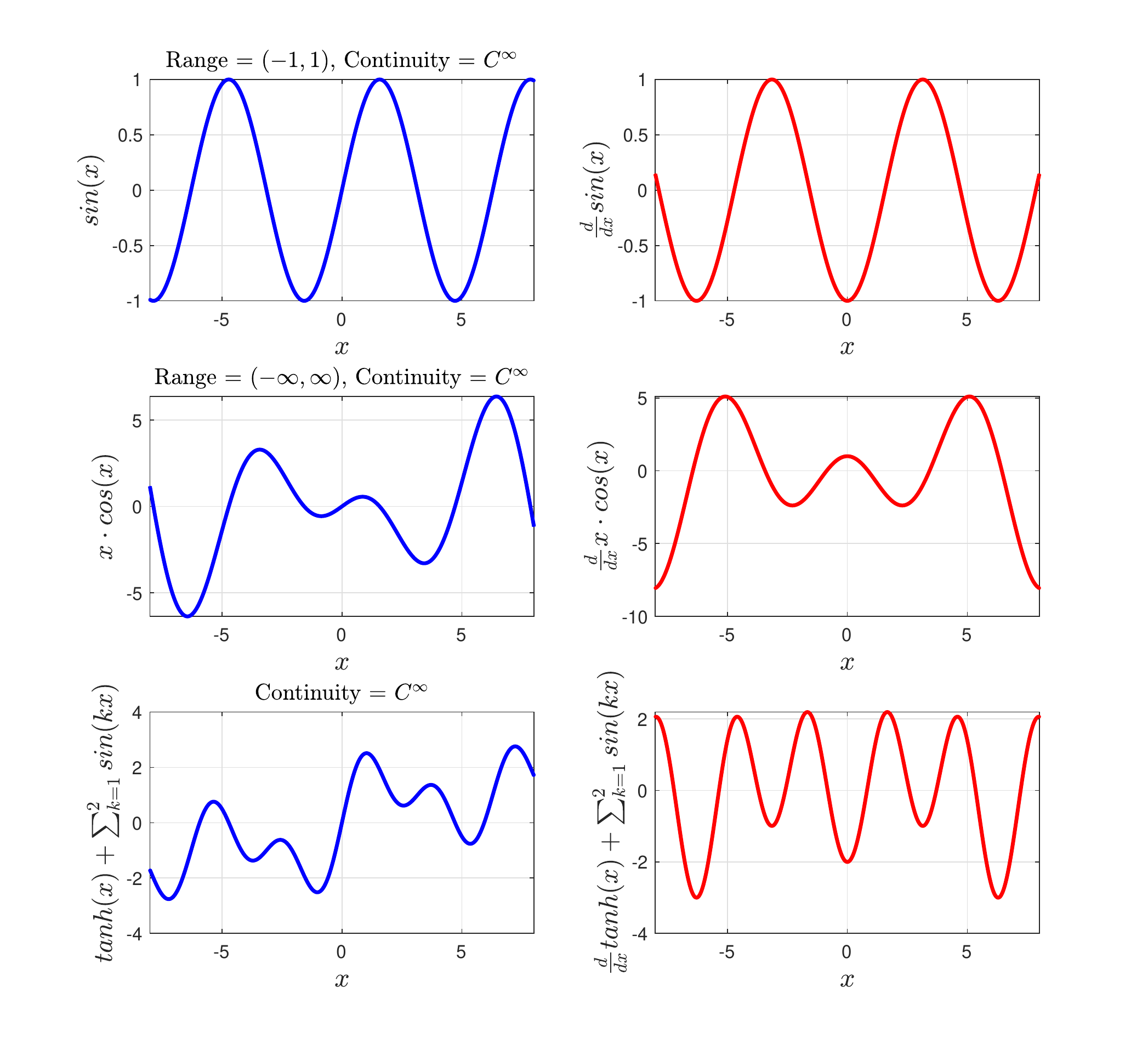}
\caption{The oscillatory activation functions (first column), and their derivatives (second column).}
\label{fig:AF_Osc}
\end{figure}
Figure \ref{fig:AF_Osc} shows various oscillatory activation functions along with their derivatives. Recently, Jagtap et al. \cite{jagtap2022deep} proposed the Rowdy activation function, where the oscillatory noise is injected over the monotonic base activation function with adaptable parameters, thereby making them oscillatory as the optimization process starts. This adaptively creates multiple hypersurfaces to better learn the data set.

\subsection{Non-standard activation functions}
This section will cover the Maxout unit and the Softmax functions, which are two non-standard activation functions. 
\begin{itemize}
\item Maxout \cite{goodfellow2013maxout} : The Maxout Unit is a piece-wise linear function that gives the maximum of the inputs, and it is designed to be used in conjunction with dropout \cite{srivastava2014dropout}. It generalizes the ReLU and leaky ReLU activation functions.
Given the units input $x\in \mathbb{R}^d$ the activation of a maxout unit is computed by first computing $k$ linear feature mappings $z \in \mathbb{R}^k$ where
$$z_i = w_i x + b_i,$$
and $w_i, b_i$ are weights and biases, whereas $k$ is the number of linear sub-units combined by one maxout unit. Later, the output $h_{\text{maxout}}$
of the maxout hidden unit is given as the maximum over the $k$ feature mappings:
$$h_{\text{maxout}} (x) = \text{max}[z_1 , . . . , z_k ].$$

Maxout's success can be partly attributed to the fact that it supports the optimization process by preventing units from remaining idle; a result of the rectified linear unit's thresholding. The activation function of the maxout unit, on the other hand, can be viewed as executing a pooling operation across a subspace of k linear feature mappings (referred to as subspace pooling in the following).
Each maxout unit is somewhat invariant to changes in its input as a result of this subspace pooling procedure. In \cite{springenberg2013improving}, Springenberg and Riedmiller proposed a stochastic generalization of the maxout unit (Probabilistic Maxout Unit) that improves each unit's subspace pooling operation while preserving its desired qualities. They first defined the probability for each of
the $k$ linear units in the subspace as follows:
$$ p_i = \frac{e^{\lambda z_i}}{\sum_{j=1}^k e^{\lambda z_j}},$$
where $\lambda$ is a chosen hyperparameter controlling the variance of the distribution. The activation $h_{\text{probout}} (x)$ is then sampled as
$$ h_{\text{probout}} (x) = z_i, ~~~\text{where}~ i \sim \text{Multinomial}\{p_1 , \ldots , p_k \}.$$
As $\lambda \rightarrow \infty$, above equation reduces to maxout unit.

  \item Softmax function : Also called as Softargmax function \cite{goodfellow2016deep} or the folding activation function or the normalized exponential function \cite{bishop2006pattern} is a generalization of logistic function in high dimensions. It normalizes the output and divides it by its sum, which forms a probability distribution.
  The standard softmax function $\Phi : \mathbb{R}^K \rightarrow (0,1)^{K}$ is defined for $K>1$ as

  $$\Phi (\mathbf {z} )_{i}={\frac {e^{z_{i}}}{\sum _{j=1}^{K}e^{z_{j}}}}\ \ \ \ {\text{ for }}i=1,\dotsc ,K $$
 and $\mathbf {z} =(z_{1},\dotsc,z_{K})\in \mathbb {R} ^{K}$. In other words, it applies the standard exponential function to each element $z_i$ of the input vector $z$ and normalizes these values by dividing them by the sum of all these exponentials; this normalization ensures that the sum of the components of the output vector $\Phi (z)$ is 1.  
\end{itemize}

\section{Complex-valued activation functions} 
The complex-valued neural network (CVNN) is a very efficient and powerful modeling tool for domains involving data in the complex number form.
Due to its suitability, the CVNN is an attractive model for researchers in various fields such as remote sensing, acoustics, opto-electronics, image processing, quantum neural devices, robotics, bioinformatics, etc.; see Hirose \cite{hirose2012complex} for more examples. 
Choosing a suitable complex activation function for CVNN is a difficult task, as a bounded complex activation function that is also complex-differentiable is not feasible. This is due to Liouville's theorem, which states that the only complex-valued functions that are bounded and analytic everywhere are constant functions. 

For CVNN there are various complex activation functions proposed in the literature. In the earlier work of Aizenberg et al. \cite{aizenberg1973multivalued}, the authors proposed the concept of \textit{multi-valued neuron} that uses the activation function $$\Phi(z) = \text{exp}(i 2\pi j/k), ~~\text{If}~ 2\pi j/k \leq \text{arg}(z) < 2\pi(j+1)/k,$$ which divides the complex plain into $k$ equal sectors thereby maps the entire complex plane onto the unit circle. In \cite{jankowski1996complex} Jankowski et al. proposed the complex-sinum function, which makes use of the multivalued threshold logic technique.
The most approaches to design CVNN preferred bounded but non-analytic functions, also called \textit{split activation function} \cite{nitta1997extension}, where real-valued activation functions are applied separately to real and imaginary part. 
Leung and Haykin \cite{leung1991complex} used sigmoid function as $\Phi(z) = 1/(1+e^{-z})$, where $z$ is complex number. Later, the following sigmoid function 
$$\Phi(z) = \frac{1}{1+e^{\text{Re}(z)}} + \frac{1}{1+e^{\text{Im}(z)}}$$
was proposed by Birx and Pipenberg \cite{birx1992chaotic}, as well as Benvenuto and Piazza \cite{benvenuto1992complex}.
The real and imaginary types of hyperbolic tangent activation function were proposed by Kechriotis and Monalakos \cite{kechriotis1994training}; see also, Kinouchi and Hagiwara \cite{kinouchi1995learning}.
Another class of split activation function for CVNN is phase-amplitude functions. Noest \cite{noest1988associative} proposed $\Phi(z) = z/|z|$, Georgious and Koutsougeras \cite{georgiou1992complex} proposed the phase-amplitude function $\Phi(z) =  \frac{z}{c +\frac{1}{r} |z|}$. These types of phase-amplitude split activation functions termed as \textit{phasor networks} by Noest \cite{noest1988associative}. Hirose \cite{hirose1994applications} proposed the following activation function $\Phi(z) = tanh(|z|) ~e^{i\text{arg}(z)}$, see also \cite{hirose1992continuous, hirose2012generalization}. 
An alternative approach where analytic and bounded (almost everywhere) fully-complex activation functions with a set of singular points were proposed by Kim and Adali \cite{kim2002fully}.
With such activation, there is a need for careful scaling of inputs and initial weights to avoid singular points during the network training. 
Apart from these complex versions of conventional activation functions, other complex activation functions have been proposed. 

A different approach to chose activation functions using conformal mappings was presented by Clarke \cite{clarke1990generalization}. 
Kuroe and Taniguchi \cite{kuroe2005models} proposed the following activation function
$$\Phi(z) =  \frac{\text{Re}(z)}{c +\frac{1}{r} |\text{Re}(z)|} + i \frac{\text{Im}(z)}{c +\frac{1}{r} |\text{Im}(z)|}.$$
A M{\"o}bius transformation based activation was  used in more general real-valued neural networks by Mandic \cite{mandic2000use}. 
A M{\"o}bius transformation based activation function were also proposed by {\"O}zdemir et al. \cite{ozdemir2011complex} as
$$\Phi(z) = \frac{az+b}{cz+d},$$
where $a,b,c,d$ are complex numbers, and $ad-cb = 1$. It is a conformal mapping of the complex plane, which is also known as bilinear transformation. In the recent years, complex-valued ReLU activation was proposed by Guberman \cite{guberman2016complex} $$ \Phi(z) = \begin{cases}
                                                                                             z & \text{If Re($z$), Img($z$)} \geq 0 \\ 0 &\text{Otherwise,}
                                                                                            \end{cases}
$$ and a modified ReLU was proposed by Arjovsky et al. \cite{arjovsky2016unitary}. 
In \cite{virtue2017better}, Virtue et al. proposed the \textit{cardioid} activation function as 
$$\Phi(z) = \frac{1}{2}(1+\cos(\angle z))z,$$
and it is used for \textit{magnetic resonance imaging (MRI)} fingerprinting.
The cardioid activation function is a complex extension of the ReLU that is phase-sensitive. 
The complex-valued Kernel activation function was proposed by Scardapane et al. \cite{scardapane2018complex}.

\section{Quantized activation functions}
The \textit{quantized neural network} (QNN) has recently attracted researchers around the world, see \cite{courbariaux2015binaryconnect, rastegari2016xnor, zhou2017incremental}. The aim of quantization is to compact the neural network model in order to improve its efficiency without degrading the performance of the model. Both forward and back-propagation processes can be executed with quantized neural networks using bitwise operations rather than floating-point operations. Weights, activations, and gradients are the three components of a neural network that may be quantified. The purposes for quantifying these components, as well as the methodologies used to quantify them, varied slightly. The size of the model can be substantially reduced by using quantized weights and activations. Moreover, with quantized gradients, the communication costs can be greatly reduced in a distributed training environment. Quantized activations can be used to replace inner-products with binary operations, allowing for faster network training.
By eliminating full-precision activations, a substantial amount of memory can be saved. The \textit{spiking neural network} (SNN) is another type of network where activation levels are quantized into temporally sparse, one-bit values, also called as '\textit{spike events}', which additionally converts the sum over weight-activity products into a simple addition of weights (one weight for each spike).

The activations were quantized to 8 bits by Vanhoucke et al. in \cite{vanhoucke2011improving}.
In particular, they quantized the activations after training the network using a sigmoid function that confines the activations to the range [0, 1]. 
In \cite{courbariaux2015binaryconnect, rastegari2016xnor, zhou2016dorefa} the authors used the binary activation as
$$ \Phi(z) = \text{Sign}(x) = \begin{cases}
                               +1 & x \geq 0, \\ -1 & x<0.
                              \end{cases}
$$
To approximate the ReLU unit, Cai et al. \cite{cai2017deep} proposed a half-wave Gaussian quantizer. They employed a half-wave Gaussian quantization function in the forward approximation.  
$$ \Phi(z) =  \begin{cases}
                               q_i & x \in (t_i, t_{i+1}], \\ 0 & x\leq 0.
                              \end{cases}$$
\begin{figure}[!h]
\centering
\includegraphics[ scale=0.72]{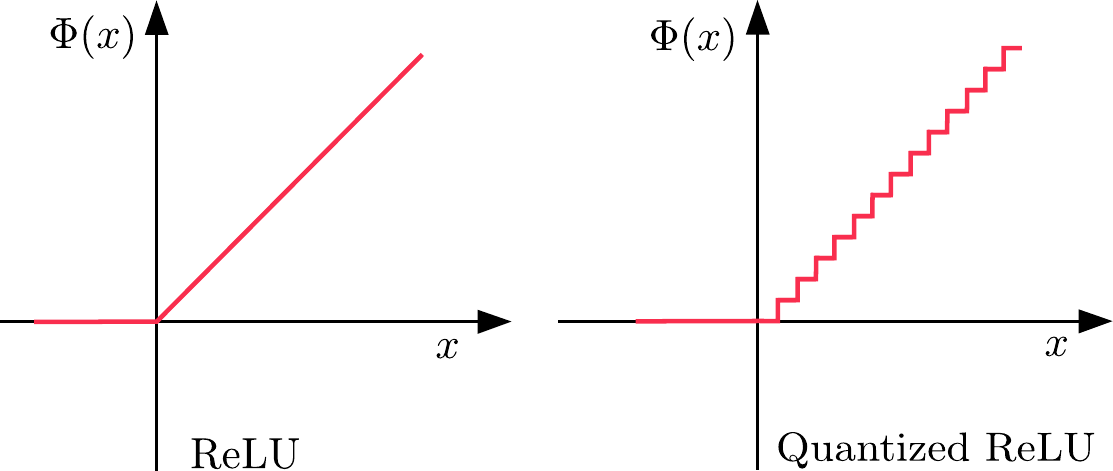}
\caption{The expected (left) and real (right) ReLU function in low-precision networks.}
\label{fig:AF_qaf}
\end{figure}
In the recent work by Anitha et al. \cite{anitha2021hyperbolic} a quantized complex-valued activation function was proposed.
The quantized activations are also proved to be successful against the adversarial examples, see Rakin et al. \cite{rakin2018defend}. Mishra et al. \cite{mishra2017wrpn} presented wide reduced-precision networks (WRPN) to quantize activation and weights. They further found that activations take up more memory than weights.
To compensate for the loss of precision due to quantization, they implemented an approach that increased the number of filters in each layer.  

One well-known difficulty with quantized activation is that it causes a gradient mismatch problem, see Lin and Talthi \cite{lin2016overcoming}. As an example, figure \ref{fig:AF_qaf} shows an expected and real ReLU activation function in low-precision networks. In a fixed point network, the effective activation function is a non-differentiable function. The \textit{gradient mismatch} problem is the result of this discrepancy between the assumed and real activation functions. 
\begin{figure}[!h]
\centering
\includegraphics[ scale=0.53]{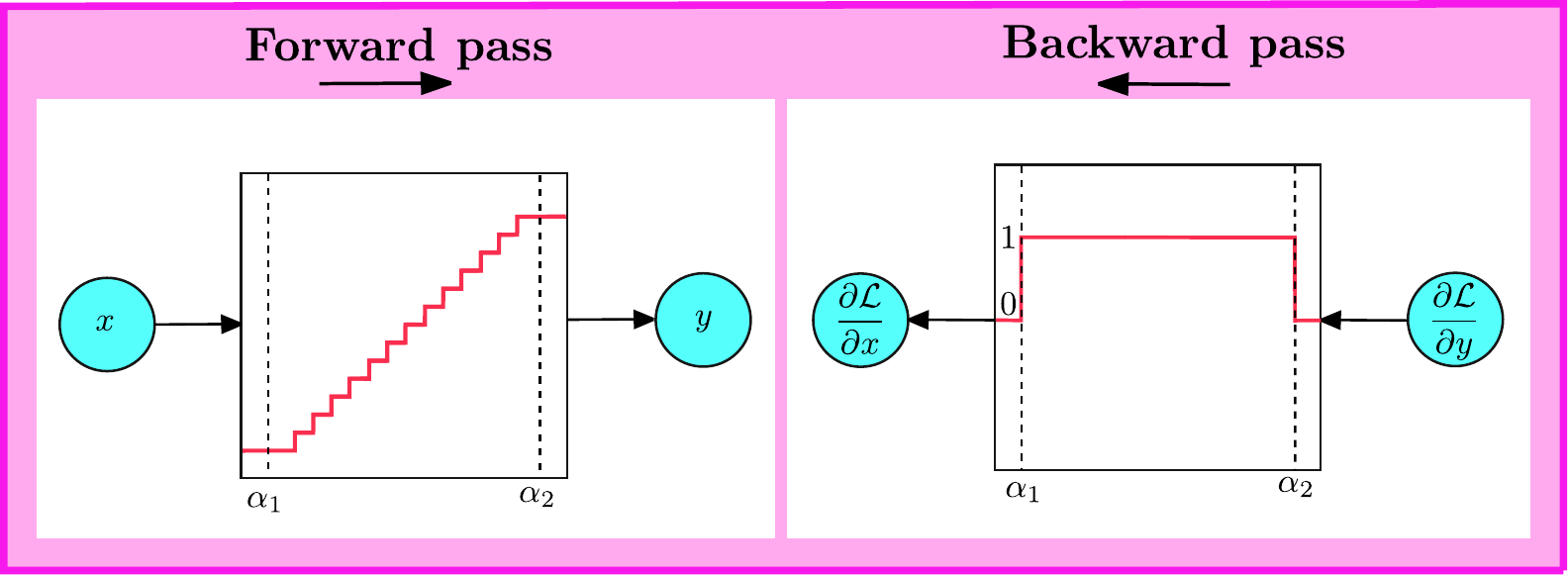}
\caption{Schematic representation of straight through estimation (STE) derivative approximation}
\label{fig:AF_qaf2}
\end{figure}
In \cite{bengio2013estimating}, Bengio et al. proposed the straight through estimation (STE) derivative approximation that works well for quantized activation function. As an example, they treated the quantized ReLU activation as the identity function in the clipping range $[\alpha_1, \alpha_2]$, and constant outside, which gives the unity derivative in the clipping range and zero outside. Figure \ref{fig:AF_qaf2} shows schematic representation of STE derivative approximation.

\section{A quest towards an optimal activation function} 
Which activation function should we use? This is one of the most basic and meaningful questions that could be posed. As discussed earlier, there is no rule of thumb for choosing the optimal activation function, which strongly depends on the problem under consideration. This motivates us to ask another meaningful question: do we need an activation function that adapts itself as per the requirements, thereby avoiding the local minima by changing the loss landscape dynamically? In this way, the adaptive activation functions can beat any standard (fixed) activation function of the same type. Despite having various activation functions, the quest for having the best activation function has driven researchers to propose an activation function that is adaptive, also called adaptable, tunable, or trainable activations.
In recent years, this has triggered a surge in papers on adaptive activation functions. The adaptive nature of the activation can be injected in many ways. For example, one can introduce parameters that adapt accordingly, or another way is to use ensemble activation functions from the pre-defined set of functions, which performs better than single activation. This section discusses various adaptive activation functions that perform better than their fixed counterparts.

\subsection{Parametric Activation Functions}
In the literature, various adaptive activation functions have been proposed. In their earlier work, Chen and Chang \cite{chen1996feedforward} proposed the generalized hyperbolic tangent function parameterized by two additional positive scalar values $a$ and $b$ as
$$\Phi(x) = \frac{a (1 - exp (-bx))}{1 + exp (-bx)},$$
where the parameters $a, b$ are initialized randomly and then adapted independently for every neuron.
Vecci et al. \cite{vecci1998learning} suggested a new architecture based on adaptive activation functions that use Catmull-Rom cubic splines. Trentin \cite{trentin2001networks} presented empirical evidence that learning the amplitude for each neuron is preferable instead of having unit amplitude for all activation functions (either in terms of generalization error or speed of convergence). Goh et al. \cite{goh2003recurrent} proposed a trainable amplitude activation function. Chandra and Singh \cite{chandra2004activation} proposed the activation function adapting algorithm for sigmoidal feed-forward neural network training. Eisenach et al. \cite{eisenach2016nonparametrically} proposed parametrically learning activation functions. Babu and Edla \cite{naresh2017new} proposed the algebraic activation functions. Ramchandran et al. \cite{ramachandran2017searching} proposed the \textit{Swish} activation function  defined as:
$$ \Phi(x) = x \cdot \text{Sigmoid}(\beta \cdot x)$$
where $\beta$ is the tuning parameter. Like ReLU, Swish is unbounded above but bounded below. The Swish is shown to give faster convergence and better generalization for many test problems.
In \cite{alcaide2018swish}, Alcaide et al. proposed the E-swish activation function as $$\Phi(x) = \beta x \cdot \text{Sigmoid}(x).$$ Mercioni et al. \cite{mercioni2020p} proposed the trainable parameter based swish (p-swish) activation that can give more flexibility than the original swish activation.
Ertu{\u{g}}rul \cite{ertuugrul2018novel} proposed the trained activation function. Choi et al. \cite{choi2018pact} proposed PArameterized Clipping acTivation (PACT) that uses an activation clipping parameter $\alpha$ that is optimized during training to find the right quantization scale.
Apicella et al. \cite{apicella2019simple} proposed an efficient architecture for trainable activation functions. 
\begin{figure*}
\centering
\includegraphics[trim=1cm 1cm 1cm 1cm, scale=0.6]{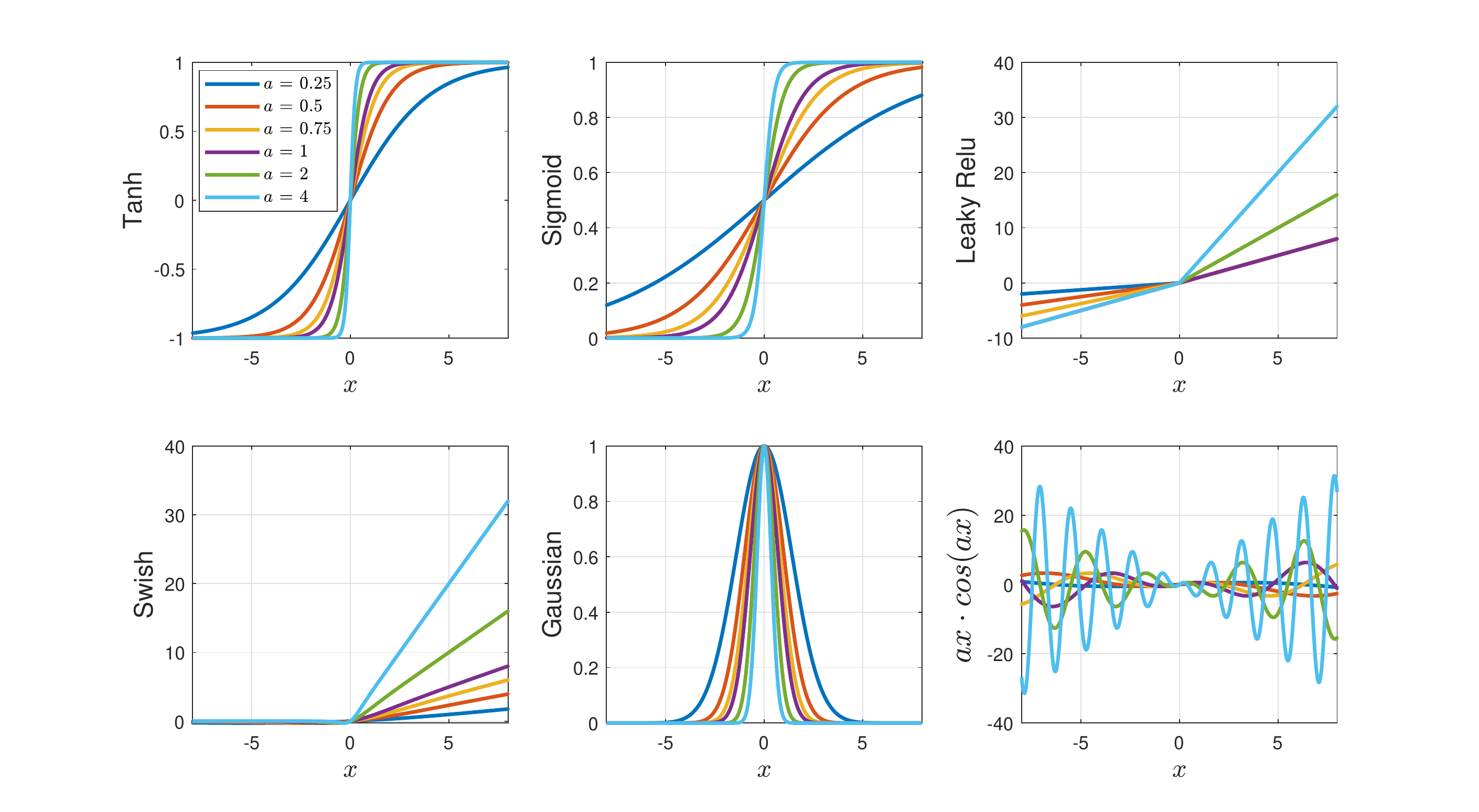}
\caption{Adaptive activation functions for different values of adaptive parameter $a$.}
\label{fig:AF_adap}
\end{figure*}
Jagtap et al. \cite{jagtap2020adaptive} proposed globally adaptive activation function where they introduced a single slope parameter $a$ in the activation function as
$$ \Phi(na(wx+b)),$$ where $a$ is a trainable parameter and  $n\geq 1$ is pre-defined scaling factor. They initialized $na = 1, \forall n\geq 1$. Figure \ref{fig:AF_adap} shows these globally adaptive activation functions for different values of $a$. Based on this idea, the layer-wise and neuron-wise locally adaptive activation functions were proposed \cite{jagtap2020locally} that can be trained faster. The main idea is to introduce trainable parameters for every hidden-layer (layer-wise adaptive activation functions) as well as for every neuron in each layer (neuron-wise adaptive activation functions).
Along with these locally adaptive activations, the additional slope recovery term is added in the activation function, which is given by
$$ S(a) = \begin{cases}
           \frac{1}{1/(D-1) \sum_{k=1}^{D-1} \text{exp}(a^k)} & \text{For layer-wise activation} \\ \frac{1}{1/(D-1) \sum_{k=1}^{D-1} \text{exp}\left(\frac{\sum_{i=1}^{Nk }a_i^k}{N_k}\right)} & \text{For neuron-wise activation},
          \end{cases}
$$
where $D$ is the depth of the network, $N_k$ is the number of neurons in the $k^{th}$ hidden-layer.  The authors in \cite{nader2020searching} proposed self-adaptive evolutionary algorithms for searching new activation functions. 
The \textit{Soft-Root-Sign} activation function  \cite{zhou2020soft} is defined as $$\Phi(x) = \frac{x}{\frac{x}{\alpha} + \text{exp}(-x/\beta)},$$ that gives range from $\left[\frac{\alpha \beta}{\beta-\alpha \times \text{exp}}, \alpha \right]$. Both $\alpha$ and $\beta$ are a pair of trainable non-negative parameters. Pratama and Kang \cite{pratama2021trainable} proposed trainable neural networks. Universal activation functions are proposed by  Yuen et al. in \cite{yuen2021universal} as 
$$\Phi(x) = \text{ln}(1+ e^{A(x+B)+Cx^2}) - \text{ln}(1+ e^{D(x-B)}) + E,$$
where $A, B$ and $E$ controls the slope, horizontal shift and vertical shift, respectively. The parameter $D$ approximates the slope of leaky ReLU, and  $C$ introduces additional degrees of freedom.

Recently, Bingham and Miikkulainen \cite{bingham2022discovering} discovered the parametric activation function using evolutionary search, where they used gradient descent to optimize its parameters during the learning process. In \cite{lapid2022evolution} Lapid and Sipper proposed a three-population, coevolutionary algorithm to evolve an activation function.

\subsubsection{Adaptive family of rectifier and exponential units}
In the literature, many adaptive families of rectifiers and exapnential units are proposed. This section gives a brief introduction to these activation functions.
\begin{itemize}
 \item \textbf{Parametric ReLU \cite{he2015delving}}: The parametric ReLU (PReLU) is similar to the leaky ReLU, and it is defined as
    $$ \Phi(x) = \begin{cases}
               \alpha x & \text{for}~x \leq 0 \\ x & \text{for}~ x > 0.
              \end{cases}
$$
In parametric ReLU, $\alpha$ is the learning parameter, which is learned during the optimization process.
Both, leaky and parametric ReLU still face the problem of exploding gradients.

 \item \textbf{S-Shaped ReLU} : Abbreviated as SReLU \cite{jin2016deep} is  defined as a combination of three linear functions, which perform a mapping $\mathbb{R} \rightarrow \mathbb{R}$ with the following formulation
$$ \Phi(x_i) =  \begin{cases}
  t_i^r + a_i^r(x_i-t_i^r), & x_i \geq t_i^r, \\ x_i, & t_i^r > x_i> t_i^l, \\ t_i^l + a_i^l(x_i-t_i^l), & x_i \leq t_i^l,
 \end{cases}$$
where $t_i^r, t_i^l, a_i^r, a_i^l$ are four learnable parameters. The subscript $i$
indicates that we allow SReLU to vary in different channels.

 \item \textbf{Parametric ELU} : The parametric ELU \cite{shah2016deep} was proposed in order to remove the need to specify the parameter $\alpha$, which can be learned during the training to get the proper activation shape at every CNN layer.
Klambauer et al. \cite{klambauer2017self} proposed Scaled exponential linear unit (SELU)
      $$ \Phi(x) = \lambda \begin{cases}
               \alpha (\text{exp}(x)-1) & \text{for}~x \leq 0 \\ x & \text{for}~ x > 0.
              \end{cases}
$$

\item \textbf{Parametric Tanh Linear Unit (P-TELU) \cite{duggal2017p}}: It is defined as
    $$ \Phi(x) = \begin{cases}
               x & \text{for}~x \geq 0 \\ \alpha~\tanh(\beta x) & \text{for}~ x < 0,
              \end{cases}
$$ which has range from [$-\alpha, \infty$), and both $\alpha$ and $\beta$ are trainable parameters. 

 \item \textbf{Continuously Differentiable ELU} : The Continuously Differentiable ELU \cite{barron2017continuously} is simply the ELU activation where the negative values have been modified to ensure that the derivative exists (and equal to 1) at $x = 0$ for all values of $\alpha$. The Continuously Differentiable ELU is defined as
       $$ \Phi(x) =  \begin{cases}
               x & \text{for}~x \geq  0 \\ \alpha (\text{exp}(x/\alpha)-1) & \text{Otherwise},
              \end{cases}
$$ where $\alpha$ is tunable parameter.

\item \textbf{Flexible ReLU \cite{qiu2018frelu}}: The Flexible ReLU, or FReLU, is defined as
$$ \Phi(x) =  ReLU(x) + b,$$
which has a range $[b, \infty)$. The FReLU captures the negative values with a rectified point.
\item \textbf{Paired ReLU \cite{tang2018joint}:}: The Paired ReLU is defined as
$$ \Phi(x) = [\text{max}(sx-\theta,0),\text{max}(s_px-\theta_p,0)],$$
where $s$ and $s_p$ represents scale parameters, which are initialized with the values of $0.5$ and -0.5, respectively . $\theta$ and $\theta_p$ are a pair of trainable thresholds.

  \item \textbf{Multiple Parametric ELU} : The multiple parametric ELU \cite{li2018improving} is given as
         $$ \Phi(x) =  \begin{cases}
                x & \text{for}~x >  0 \\ \alpha (\text{exp}(\beta x )-1) & \text{for}~ x \leq 0.
              \end{cases}
$$ Here, $\beta$ is greater than zero.
   \item \textbf{Parametric Rectified Exponential Unit} : The parametric rectified EU \cite{ying2019rectified} is defined as
   
            $$ \Phi(x) =  \begin{cases}
                \alpha x & \text{for}~x >  0 \\ \alpha x ~\text{exp}(\beta x ) & \text{for}~ x \leq 0,
              \end{cases}
$$
    \item \textbf{Fast ELU} : The Fast ELU \cite{qiumei2019improved} is given as
                $$ \Phi(x) = \begin{cases}
                x & \text{for}~x >  0 \\ \alpha (\text{exp}(x/\text{ln}(2) )-1) & \text{for}~ x \leq 0.
              \end{cases}
$$ The curve trend with fast ELU is consistent with the ELU due to the fast approximation achieved, ensuring that the fast ELU does not alter the original ELU's accuracy advantage. 

\item \textbf{Multi-bin Trainable Linear Units (MTLU) \cite{gu2019fast}}: The MTLU is given as

    $$ \Phi(x) = \begin{cases}
               a_0 x+ b_0 & \text{for}~x \leq c_0, \\ a_1 x+ b_1 & \text{for}~ c_0 < x \leq c_1, \\ \cdots \\ a_r x+ b_r & \text{for}~ c_{r-1} < x, 
              \end{cases}
$$ with range ($-\infty,\infty$). The MTLU considers different bins with different ranges of hyperparameters. All the parameters $a_0, a_1,\cdots, a_r, b_0, b_1,\cdots, b_r, c_0, c_1,\cdots, c_r$ are trainable.

\item \textbf{Mexican ReLU}: The Mexican ReLU \cite{maguolo2021ensemble} uses Mexican hat type function and it is defined as:

$$\Phi(x) = \text{PReLU}(x) + \sum_{i=1}^K c_i ~\text{max}(\lambda_i-|x-a_i|, 0),$$
where $c_i$ are tunable parameter and $\lambda_i, a_i$ are real numbers.
 
\item \textbf{Lipschitz ReLU \cite{basirat2020relu}}: The L* ReLU is defined as
    $$ \Phi(x) = \begin{cases}
               \text{max}(\phi(x),0) & \text{for}~x \geq 0 \\ \text{max}(\eta(x),0) & \text{for}~ x < 0,
              \end{cases}
$$ where $\phi(x)$ and $\eta(x)$ are nonlinear functions.

  \item \textbf{Parametric Deformable ELU} : First defined in \cite{cheng2020parametric}, the parametric deformable ELU is given as
  
           $$ \Phi(x) =  \begin{cases}
                x & \text{for}~x >  0 \\ \alpha [(1+(1-t)x)^{1/(1-t)} - 1] & \text{for}~ x \leq 0,
              \end{cases}
$$ with range $(-\alpha, \infty)$.
   \item \textbf{Elastic ELU} : The Elastic ELU \cite{kim2020elastic} is designed to utilize the properties of both ELU and ReLU together, and it is defined as
   $$ \Phi(x) =  \begin{cases}
                kx & \text{for}~x >  0 \\ \alpha (\text{exp}(\beta x )-1) & \text{for}~ x \leq 0,
              \end{cases}
$$ where both $\alpha$ and $\beta$ are tunable parameters.

\end{itemize}

\subsection{Stochastic/probabilistic adaptive activation functions} 
The stochastic/probabilistic approach is another way to introduce an adaptive activation function. In \cite{gulcehre2016noisy}, Gulcehre et al. proposed a noisy activation function where a structured and bounded noise-like effect is added to allow the optimizer to exploit more and learn faster. This is particularly effective for the activation functions, which strongly saturate for large values of their inputs. In particular, they learn the level of injected noise in the saturated regime of the activation function.
They consider the noisy activation functions of the following form:
$$\Phi_{\text{noisy}}(x) = \Phi(x) + s$$
where $s = \mu + \sigma \xi$. Here $\xi$ is an \textit{independent and identically distributed} random variable drawn from generating distribution, and the parameters $\mu$ and $\sigma$ are used to generate a location scale family from $\xi$.
When the unit saturates, they pin its output to the threshold value $t$, and the noise is added.
The type of noise $\xi$ and the values of $\mu$ and $\sigma$, which can be chosen as functions of $x$ to allow some gradients to propagate even in the saturating regime, determine the exact behavior of the method. 

\begin{figure}[!h]
\centering
\centering
\includegraphics[scale=0.9]{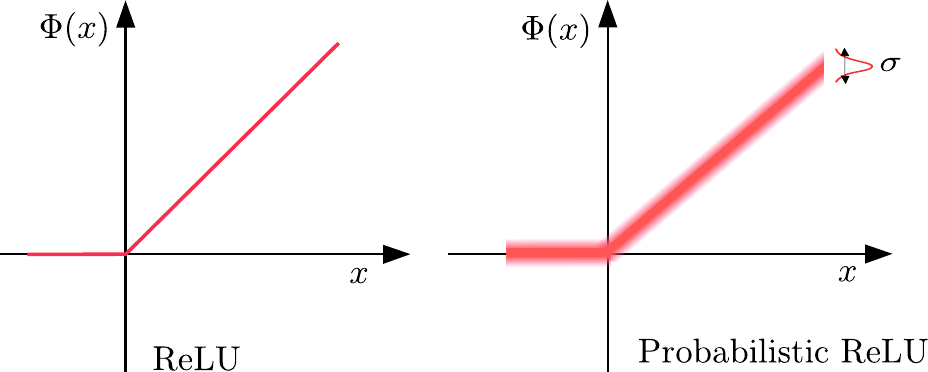}
\caption{ReLU (left) and Probabilistic ReLU (right) activation functions.}
\label{fig:AF_PROB}
\end{figure}

Urban et al. \cite{urban2017gaussian} proposed Gaussian process-based stochastic activation functions. Based on a similar idea by Gulcehre et al. \cite{gulcehre2016noisy}, Shridhar \cite{shridhar2019probact} proposed the \textit{probabilistic activation functions (ProbAct)}, which are not only trainable but also stochastic in nature, see Figure \ref{fig:AF_PROB}. The ProbAct is defined as:
$$\Phi(x) = \mu(x) + z,$$
where $\mu(x)$ is a static or learnable mean, say, $\mu(x) = max(0, x)$ for static ReLU, and
the perturbation term $z = \sigma \epsilon$, where perturbation parameter $\sigma$ is a fixed or trainable value which specifies the range of stochastic
perturbation and $\epsilon$ is a random value sampled from a normal distribution $\mathcal{N}(0, 1)$. Because the ProbAct generalizes the ReLU activation, it has the same drawbacks as the ReLU activation.

The \textit{Rand Softplus} activation function \cite{chen2019improving} models the stochaticity-adaptibility as $$\Phi(x) = (1-\rho) ~\text{max}(0,x) + \rho~ \text{log}(1+e^x),$$
where $\rho$ is a stochastic hyperparameter.

\subsection{Fractional adaptive activation functions}
The fractional activation function can encapsulate many existing and state-of-the-art activation functions. In the earlier study of Ivanov \cite{ivanov2018fractional} the fractional activation functions are presented. This is motivated by the potential benefits of having more tunable hyperparameters in a neural network and achieving different behaviors, which can be more suitable for some kinds of problems. In particular, the author used the following Mittag-Leffler types of functions:
$$E_{\alpha,\beta} (z) = \sum_{k=0}^{\infty} \frac{z^k}{\Gamma(\alpha k+\beta)}, z\in \mathbb{C},$$
which is the class of parametric transcendental functions that generalize the exponential function. Setting $\alpha$ and $\beta$ to unity gives the Taylor series expansion of the exponential function.
The authors proposed several fractional activation functions by replacing the exponential function with several standard activation functions such as tanh, sigmoid, logistic, etc.

\begin{figure*}[!h]
\centering
\includegraphics[trim=1cm 1cm 0cm 0cm,  scale=0.45]{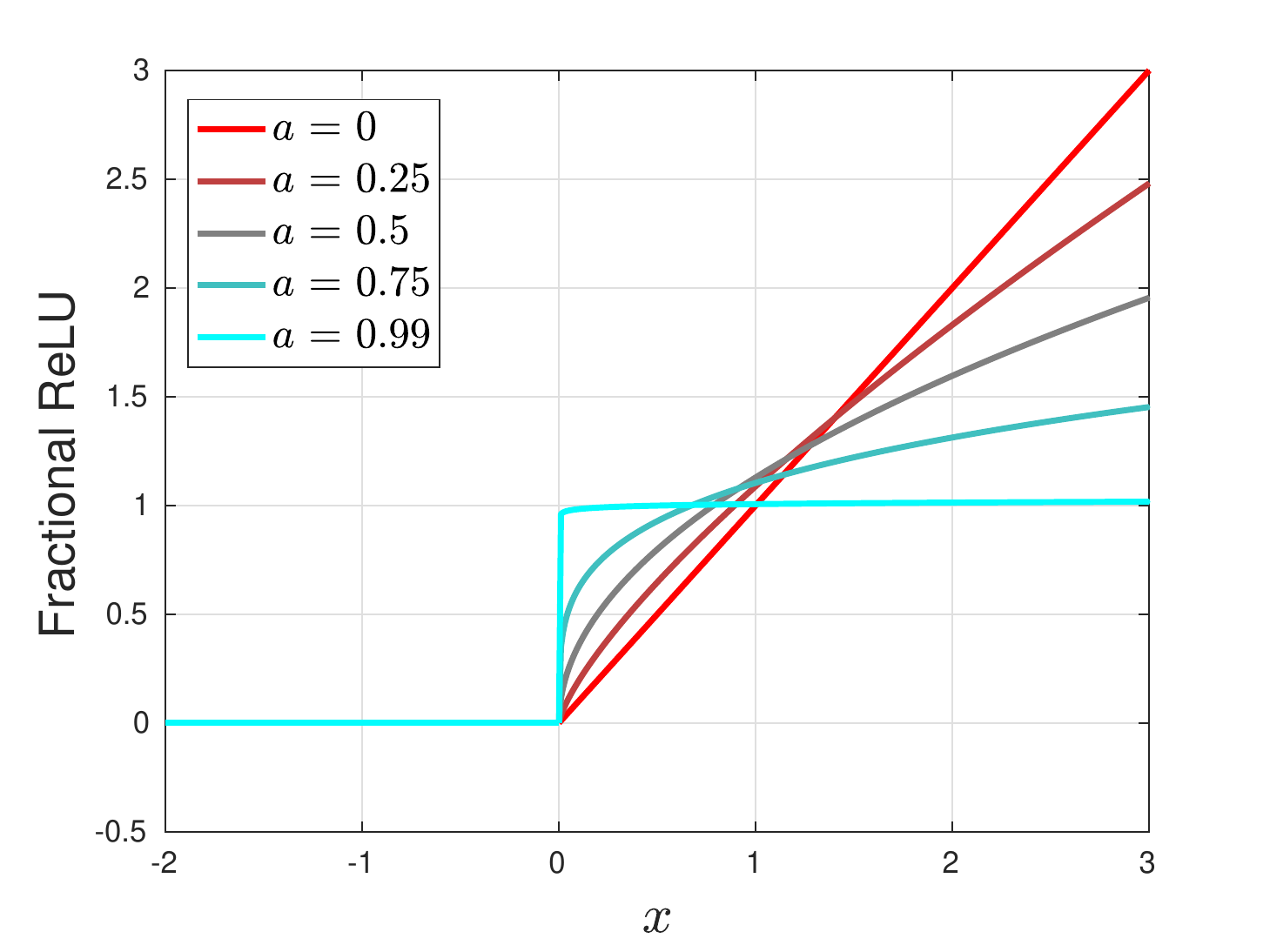}
\includegraphics[trim=1cm 1cm 0cm 0cm,   scale=0.45]{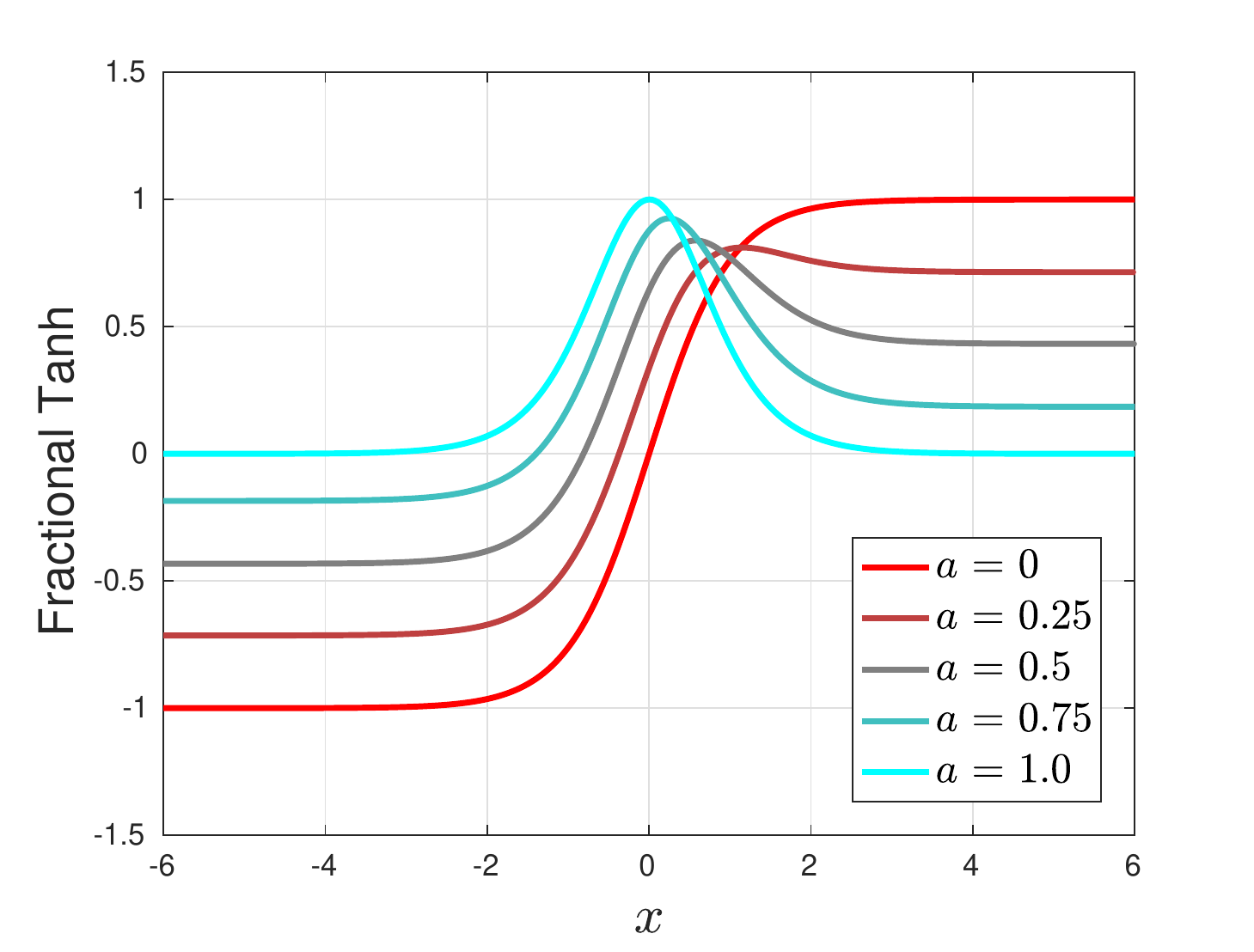}
\caption{Fractional ReLU (left), and fractional hyperbolic tangent (right) activation functions for different values of fractional derivative $a$.}
\label{fig:AF_frac}
\end{figure*}

In \cite{zamora2019adaptive}, the authors proposed activation functions leveraging the fractional calculus. In particular, they used the Gamma function ($\Gamma(z) = \int_0^{\infty} t^{z-1} e^{-t} ~dt $). The fractional ReLU is then defined as

$$\Phi(x) = \begin{cases}
D^a x & x >0, \\ 0 & x\leq 0,
\end{cases}
$$

where the fractional derivative is given by $$D^a x = \frac{\Gamma(2)}{\Gamma(2-a)} x^{1-a}.$$ Similarly, the fractional derivative of hyperbolic tangent is given by
\begin{equation*}\Phi(x) =  D^a tanh(x) = \lim_{h\rightarrow 0} \frac{1}{h^a} \sum_{n=0}^{\infty} (-1)^n \frac{\Gamma(a+1)~tanh(x-n\cdot h)}{\Gamma(n+1)\Gamma(1-n+a)}.
\end{equation*}
The fractional derivative produces a family of functions for different values of $a$. Figure  \ref{fig:AF_frac} depicts the fractional ReLU (left) and fractional hyperbolic tangent (right) functions for different $a$ values. At $a = 0.99$, the ReLU becomes a step function, whereas the hyperbolic tangent function ($a = 0$) becomes a quadratic hyperbolic secant function.

\subsection{Ensemble adaptive activation functions}
An alternative approach to adaptive activation functions is to create a combination of different activation functions. The earlier work of Xu and Zhang \cite{xu2000justification} proposes an adaptive ensemble activation function consisting of sigmoid, radial basis, and sinusoid functions. In \cite{ismail2013predictions}, Ismail et al. select the most suitable activation function as a discrete optimization problem that involves generating various combinations of functions. The proposed activation function is expressed as 
$$\Phi(x) = \sum_{i=1}^n k_i~ c_i~ \psi_i(x),$$
where $n$ is the number of the sub-functions represented by $(\psi_i)$ in the activation function, $c_i$ is an adaptive coefficient, and $k_i$ is a binary number.

Chen \cite{chen2016combinatorially} used multiple activation functions for each neuron for the problems related to stochastic control. Jin et al. \cite{jin2016deep} proposed the combination of a set of linear functions with open parameters. Agostinelli et al. \cite{agostinelli2014learning} constructed the activation functions during network training. They used the following general framework:
$$\Phi(x) = ~\text{max}(0,x) + \sum_{p = 1}^P a_i^p ~\text{max}(0,-p+b_i^p),$$
where $P$ is predefined hyperparameter, $i$ is the neuron, and $a,b$ are trained variables.

Ramachandran et al \cite{ramachandran2017searching} used a reinforcement learning controller to combine preset unary and binary functions for learning new activation functions. In \cite{qian2018adaptive}, Qian et al. proposed adaptive activations in convolutional neural networks, where they focus on learning activation functions via combining basic activation functions in a data-driven way. In particular, they used ReLU and other variants of ReLU, such as Leaky ReLU and ELU combinations. In \cite{klabjan2019activation}, Klabjan and Harmon show that the ensemble activation function can be created by choosing the suitable activation functions from the predefined set $A = \{A_1, A_2,\cdots,A_m\}$. Here, the activations are combined together to get the best performance out of the neural networks. A similar approach was used by Nandi et al. in \cite{nandi2020improving}, where an ensemble of activation functions is used to improve the performance of the network. In \cite{apicella2019simple} , Apicella et al. proposed to modify a feed-forward neural network by adding Variable Activation Functions (VAF) in terms of one-hidden layer subnetworks. The previously discussed Mexican ReLU \cite{maguolo2021ensemble} also comes under the category of ensemble activation functions.

A polynomial activation function was proposed in \cite{piazza1992artificial} where they built the activation function over powers of the activation. They used
$$\Phi(x) = \sum_{i=0}^P \alpha_i x^i,$$
where $P$ is a hyper-parameter. Because a polynomial of degree $P$ can pass through all $P + 1$ points exactly, this polynomial activation function can theoretically approximate any smooth function. The main disadvantage was the global influence of coefficient $\alpha_i$, which causes the function's output to grow excessively large. Similarly, Scardapane et al. \cite{scardapane2019kafnets} proposed the kernel-based non-parametric activation function. In particular, they model each activation function in terms of a kernel expansion over a finite number of terms as
$$\Phi(x) = \sum_{i=1}^D \alpha_i \kappa(x,d_i),$$
where $\{\alpha_i\}_{i=1}^D$ are the mixing coefficients, $\{d_i\}_{i=1}^D$ are the called the dictionary elements, and $\kappa(·, ·) : \mathbb{R} \times \mathbb{R} \rightarrow \mathbb{R}$ is a one-dimensional kernel function. Further, this idea was extended with a multi-kernel approach in \cite{scardapane2019multikernel}.
On similar grounds, the adaptive blending units \cite{sutfeld2020adaptive} were proposed by S{\"u}tfeld et al. to combine a set of functions in a linear way.
In \cite{basirat2018quest}, Basirat and Roth presented a genetic algorithm based learning of activation functions where the hybrid crossover of various operators results in a hybrid activation function.

\begin{table*}[!h]
\centering
\scriptsize
\begin{tabular}{|c|c|c|c|c|c|c|} 
\hline
& & & & & & \\
References & \textbf{Parametric}& \textbf{Ensemble} & Stochastic/ & Fractional & Complex-Valued & Quantized \\
 & &  & Probabilistic &  &  &  \\
\hline
& & & & & & \\
Chen and Chang \cite{chen1996feedforward}, Vecci \cite{vecci1998learning}, Trentin \cite{trentin2001networks} & & & & & & \\  
Goh \cite{goh2003recurrent}, Chandra and Singh \cite{chandra2004activation},  Shah et al. \cite{shah2016deep},& & & & & & \\ 
 Jin et al.\cite{jin2016deep}, Shah \cite{shah2016deep}, Jin \cite{jin2016deep}  & \cmark  & \xmark  & \xmark & \xmark & \xmark  & \xmark \\
 He et al. \cite{he2015delving}, Barron \cite{barron2017continuously}, Duggal and Gupta \cite{duggal2017p}  & & & & & & \\ 
 Li et al. \cite{li2018improving},  Tang et al.\cite{tang2018joint}, Qiu et al. \cite{qiu2018frelu} & & & & & & \\ 
 Grelsson and Felsberg \cite{grelsson2018improved},   Ying et al. \cite{ying2019rectified}, Qiumei et al. \cite{qiumei2019improved} & & & & & & \\ 
 Gu et al. \cite{gu2019fast}, Goyal et al. \cite{goyal2019learning}, Jagtap et al. \cite{jagtap2020adaptive, jagtap2020locally}  & & & & & & \\ 
Basirat and Roth \cite{basirat2020relu}, Cheng et al. \cite{cheng2020parametric}, Kim et al. \cite{kim2020elastic} & & & & & & \\ 
 \hline
& & & & & & \\
Piazza et al. \cite{piazza1992artificial}, Agostinelli et al. \cite{agostinelli2014learning} Jagtap et al. \cite{jagtap2022deep} & \cmark  & \cmark  & \xmark & \xmark & \xmark  & \xmark\\
Scardapane et al. \cite{scardapane2019kafnets} & & & & & & \\ \hline
& & & & & & \\
Gulcehre et al. \cite{gulcehre2016noisy}, Shridhar et al. \cite{shridhar2019probact}, Urban et al.\cite{urban2017gaussian} & \cmark  & \xmark  & \cmark & \xmark & \xmark  & \xmark \\
& & & & & & \\
\hline
& & & & & & \\
Zamora et al. \cite{zamora2019adaptive}, Ivanov et al. \cite{ivanov2018fractional} & \cmark  & \xmark  & \xmark & \cmark & \xmark & \xmark \\
& & & & & & \\ \hline & & & & & & \\
Xu and Zhang \cite{xu2000justification}, Ismail et al. \cite{ismail2013predictions} & & & & & & \\ 
 Ramchandran et al. \cite{ramachandran2017searching}, Klabjan and Harmon \cite{klabjan2019activation} & \xmark  & \cmark  & \xmark & \xmark & \xmark  & \xmark\\
S{\"u}tfeld et al. \cite{sutfeld2020adaptive}, Basirat and Roth \cite{basirat2018quest} & & & & &&  \\ \hline & & & & & & \\

Nanni et al \cite{nanni2020stochastic} & \xmark  & \cmark  & \cmark & \xmark & \xmark & \xmark  \\  & & & & & & \\ \hline & & & & & & \\

Choi et al. \cite{choi2018pact}, Rakin et al. \cite{rakin2018defend} & \cmark  & \xmark  & \xmark & \xmark & \xmark & \cmark  \\  & & & & & & \\ \hline
\end{tabular}
\caption{Adaptive activation functions under various settings.}
\label{CompVCX}
\end{table*}

In recent years, Nanni et al. \cite{nanni2020stochastic} proposed stochastic selection of activation layers for convolutional neural networks. In \cite{jagtap2022deep} Jagtap et al. proposed the Kroneker Neural Networks (KNN), which is a general framework for adaptive activation functions that can generalize a class of existing feed-forward neural networks that utilize adaptive activation functions. In particular, the output of KNN is given as
$$
u^{\mathcal{K}}_{\Theta_{\mathcal{K}}}(x) = \sum_{i=1}^N c_i \left[ \sum_{k=1}^K \alpha_k^i\Phi_k(\omega_k^i(w_i^Tx + b_i))\right],
$$
where the network parameters are: $\Theta_{\mathcal{K}} = \{c_i, w_i, b_i\} \cup \{\alpha_k^i, \omega_k^i\}$ for $i = 1, \cdots, N$ and $k = 1,\cdots, K$. Different classes of adaptive activation functions are produced by varying these parameters and the hyperparameter $K$. For example,

\begin{itemize}
    \item If $K=1$, $\omega^i = \alpha^i = 1$ for all $i$,
    the Kronecker network 
    becomes a standard feed-forward network.
    \item If $K = 1$, the Kronecker network 
    becomes a feed-forward neural network with 
    layer-wise locally adaptive activation functions \cite{jagtap2020adaptive, jagtap2020locally}.
    \item If $K=2$, $\omega^i_1 = 1$, $\omega^i_2 = \omega_2$ for all $i$, 
    $\Phi_1(x) = \max\{x, 0\}$,
    and $\Phi_2(x) = \max\{-x,0\}$,
    the Kronecker network 
    becomes a feed-forward network with Parametric ReLU activation \cite{he2015delving}. 
    \item If $K=2$, $\omega^i_2 = \omega$ for all $i$, 
    $\Phi_1(x) = \max\{x, 0\}$,
    and $\Phi_2(x) = (e^x - 1)\cdot \mathbb{I}_{x \le 0}(x)$,
    the Kronecker network 
    becomes a feed-forward network with Exponential Linear Unit (ELU)  activation \cite{clevert2015fast}
    if $\omega^i_1 = 1$ for all $i$, 
    and 
    becomes a feed-forward network with Scaled Exponential Linear Unit (SELU) activation \cite{klambauer2017self} 
    if $\omega^i_1 = \omega'$ for all $i$.
    \item If $\omega^i = 1$ for all $i$ and 
    $\Phi_k(x) = x^{k-1}$ for all $k$, 
    the Kronecker network 
    becomes a feed-forward neural network with 
    self-learnable activation functions (SLAF) \cite{goyal2019learning}.
    Similarly, a FNN with smooth adaptive activation function \cite{hou2017convnets}
    can be represented by a Kronecker network.
\end{itemize}
In \cite{nanni2022comparison} Nanni et al. used the ensembling method to compare different convolutional neural network activation functions. In particular, they used the set of twenty activation functions.

Every type of adaptive activation that has been discussed in this section has pros and cons of its own. Table \ref{CompVCX} shows the adaptive activation function referenced under various settings.

\section{Performance of some fixed and adaptive activation functions for classification tasks}
This section discusses the performance of different activation functions for classification tasks. In particular, we used MNIST \cite{lecun1998gradient}, CIFAR-10, and CIFAR-100 \cite{krizhevsky2009learning} data sets for the same. The MNIST database of handwritten digits contains the training set of 60k samples, while the test set contains 10k examples. The CIFAR-10 data set contains 50k training and 10k testing images from 10 object categories, whereas the CIFAR-100 data set contains 50k training and 10k testing images from 100 object categories. In particular, for comparison, we chose some fixed as well as some adaptive activation functions that have been discussed previously.
\begin{table*}[!h]
\footnotesize
\centering
\begin{tabular}{cccc} 
\hline
 &\textbf{MNIST} & \textbf{CIFAR-10}&\textbf{CIFAR-100} \\\hline
\textbf{Sigmoid} & 97.9 \cite{pedamonti2018comparison} & 89.43 $\pm$ 0.51 (M), 85.42 $\pm$ 0.47 (V) & 61.64 $\pm$ 0.56 (M), 59.25 $\pm$ 0.45 (V)\\
\textbf{Tanh} & 98.21\cite{eisenach2016nonparametrically} & 88.19 $\pm$ 1.21 (M), 87.53 $\pm$ 0.67 (V) & 57.06 $\pm$ 2.03 (M), 62.32 $\pm$ 0.82 (V)\\ 
\textbf{Swish} & 99.75 $\pm$ 0.11 (M), 99.45 $\pm$ 0.26 (V) & 95.5 \cite{ramachandran2017searching} & 83.9 \cite{ramachandran2017searching}\\
\textbf{ReLU} & 99.1 \cite{apicella2019simple}, 99.15 \cite{scardapane2019kafnets}, 99.53 \cite{jin2016deep}  &  95.3 \cite{ramachandran2017searching}, 94.59 \cite{trottier2017parametric}, 92.27 \cite{jin2016deep} & 83.7 \cite{ramachandran2017searching}, 75.45 \cite{trottier2017parametric}, 67.25 \cite{jin2016deep}  \\
\textbf{Leaky ReLU} & 98.2 \cite{pedamonti2018comparison},99.58 \cite{jin2016deep}  & 95.6\cite{ramachandran2017searching}, 92.32 \cite{qian2018adaptive} &  83.3\cite{ramachandran2017searching}, 67.3\cite{jin2016deep}  \\
\textbf{ELU} & 98.3 \cite{pedamonti2018comparison}  & 94.4 \cite{ramachandran2017searching}, 94.01 \cite{trottier2017parametric} &  80.6 \cite{ramachandran2017searching}, 74.92\cite{trottier2017parametric}  \\
\textbf{PReLU} & 99.64 $\pm$ 0.24 (M), 99.18 $\pm$ 0.17 (V) &  92.46 $\pm$ 0.44 (M), 91.63 $\pm$ 0.31 (V) & 69.46 $\pm$ 0.74 (M), 66.53$\pm$ 0.69 (V)\\
\textbf{SELU} & 98.42 $\pm$ 0.53 (M), 99.02 $\pm$ 0.37(V) &  93.52 $\pm$ 0.63 (M), 90.53 $\pm$ 0.36 (V) & 70.42 $\pm$ 0.75 (M), 68.02$\pm$ 1.29 (V)\\
\textbf{RReLU} & 99.23 $\pm$ 0.53 (M), 99.63 $\pm$ 0.6 (V) &  90.52 $\pm$ 2.14 (M), 90.18 $\pm$ 0.91 (V) & 68.62 $\pm$ 0.42 (M), 65.32 $\pm$ 1.74 (V)\\
\textbf{GELU} & 99.72 $\pm$ 0.35 (M), 99.26 $\pm$ 0.42 (V) &  94.76 $\pm$ 0.55 (M), 92.67 $\pm$ 0.89 (V) & 71.73 $\pm$ 1.09 (M), 69.61 $\pm$ 1.53 (V)\\
\textbf{CELU}& 99.36 $\pm$ 0.68 (M), 99.37 $\pm$ 0.38 (V) &  90.26 $\pm$ 0.12 (M), 90.37 $\pm$ 0.23 (V) & 70.26 $\pm$ 1.53 (M), 68.35 $\pm$ 0.87 (V)\\
\textbf{Softplus} & 98.69 $\pm$ 0.5 (M), 97.36 $\pm$ 0.77 (V) & 94.9 \cite{ramachandran2017searching} & 83.7 \cite{ramachandran2017searching}\\
\textbf{Mish} & 97.9 \cite{pedamonti2018comparison} &  90.26 $\pm$ 0.52 (M), 86.05 $\pm$ 0.76 (V) & 68.53 $\pm$ 0.86 (M), 67.03 $\pm$ 1.39 (V)\\
\textbf{Maxout} & 99.55 \cite{goodfellow2013maxout} & 90.62 \cite{goodfellow2013maxout} & 61.43 \cite{goodfellow2013maxout}\\
\textbf{SRS} & 98.04 $\pm$ 0.97 (M), 98.06 $\pm$ 0.84 (V) &  89.35 $\pm$ 0.85 (M), 87.26 $\pm$ 1.38 (V) & 65.20 $\pm$ 1.53 (M), 63.65 $\pm$ 2.63 (V)\\
\textbf{Ensemble (\cite{klabjan2019activation})}&  99.40 \cite{klabjan2019activation} & 85.05 $\pm$ 0.28 (M), 84.96 $\pm$ 0.87 (V) & 74.20 \cite{klabjan2019activation} \\ 
\textbf{LiSHT} & 98.74 $\pm$ 0.17 (M), 98.32 $\pm$ 0.19 (V) &  90.78 $\pm$ 0.43 (M),87.74 $\pm$ 0.36 (V) & 56.35 $\pm$ 0.58 (M), 58.74 $\pm$ 0.99 (V)\\ 
\textbf{Sine} &99.10 $\pm$ 0.82 (M), 98.63 $\pm$ 0.27 (V) &  91.64 $\pm$ 0.36 (M), 90.59 $\pm$ 0.88 (V) & 80.98 $\pm$ 0.81 (M), 78.42 $\pm$ 0.67 (V)\\
\textbf{GCU (\cite{noel2021growing})} &97.80 $\pm$ 1.02 (M), 90.73 $\pm$ 0.87 (V) &  90.24 $\pm$ 1.96 (M), 89.56 $\pm$ 0.18 (V) & 77.46 $\pm$ 2.91 (M), 76.46 $\pm$ 1.61 (V)\\
\textbf{Gaussian} &98.60 $\pm$ 2.22 (M), 96.43 $\pm$ 2.57 (V) &  90.64 $\pm$ 2.76 (M), 89.59 $\pm$ 0.28 (V) & 76.96 $\pm$ 0.41 (M), 74.62 $\pm$ 2.69 (V)\\
\textbf{Adaptive sine } &99.56 $\pm$ 0.56 (M), 98.76 $\pm$ 0.12 (V) &  92.64 $\pm$ 0.36 (M), 92.72 $\pm$ 0.37 (V) &   81.15 $\pm$ 0.81   (M), 78.73 $\pm$ 0.64 (V)\\
\textbf{Adaptive tanh} &99.52 $\pm$ 0.36 (M), 99.57 $\pm$ 0.20 (V) &  91.14 $\pm$ 0.45 (M), 90.52 $\pm$ 0.82 (V) & 83.82 $\pm$ 0.83 (M), 76.21 $\pm$ 1.95 (V)\\
\textbf{Adaptive ReLU} &99.75 $\pm$ 0.25 (M), 98.43 $\pm$ 0.32 (V) &  96.89 $\pm$ 0.93 (M), 94.77 $\pm$ 1.16 (V) & $\mathbf{84.58\pm 0.42}$ (M), 80.30 $\pm$ 0.60 (V)\\
\textbf{Rowdy Sine} &99.74 $\pm$ 0.29 (M), 99.24 $\pm$ 0.43 (V) &  93.87 $\pm$ 0.78 (M), 92.82 $\pm$ 0.82 (V) & 83.73 $\pm$ 0.83 (M), 82.14 $\pm$ 0.13 (V)\\
\textbf{Rowdy tanh} &99.45 $\pm$ 0.93 (M), 98.65 $\pm$ 0.57 (V) &  93.78 $\pm$ 0.13 (M), 93.26 $\pm$ 0.42 (V) & 82.34 $\pm$ 1.01 (M), 80.97 $\pm$ 0.32 (V)\\
\textbf{Rowdy ReLU} &99.43 $\pm$ 0.68 (M), 98.23 $\pm$ 0.25 (V) &  $\mathbf{95.57 \pm 0.84}$ (M), 94.47 $\pm$ 0.92 (V) & $82.75 \pm 0.51$ (M), 79.41 $\pm$ 0.75 (V)\\
\textbf{Rowdy ELU} & $\mathbf{99.79 \pm 0.08}$ (M), 99.06 $\pm$ 1.07 (V) &  95.45 $\pm$ 0.92 (M), 93.27 $\pm$ 0.13 (V) & 84.29 $\pm$ 0.13 (M), 82.40 $\pm$ 0.81 (V)\\
\hline
\end{tabular}
\caption{Accuracy comparison of different activation functions for MNIST, CIFAR-10, and CIFAR-100 data sets. Here, we report the best accuracy from the literature without considering the architecture used to perform the experiments. Also, we used MobileNet \cite{howard2017mobilenets} (denoted by M) and \textit{VGG16} \cite{simonyan2014very} (denoted by V) architectures with different activation functions, where we report the mean and standard deviations of 10 different realizations. The \textit{Adaptive} and \textit{Rowdy} (with $K=3$) refers \cite{jagtap2020adaptive} and \cite{jagtap2022deep}, respectively. The best results are highlighted in bold.}
\label{AF_CvRNN}
\end{table*}
In this work, we report the best accuracy from the literature without considering the architecture used to perform the experiments. We also conducted a performance comparison of two models, namely, \textit{MobileNet} \cite{howard2017mobilenets} and \textit{VGG16} \cite{simonyan2014very}. Both MobileNet and VGG16 have a convolution neural net (CNN) architecture. The MobileNet is a lightweight deep neural network with better classification accuracy and fewer parameters, whereas the VGG16 has 16 deep layers.
In both cases, the learning rate is 1e-4, and the batch sizes for MNIST, CIFAR-10 and CIFAR-100 are 64, 128, and 64, respectively. During training and testing, data normalization is performed. The Adam \cite{kingma2014adam} optimizer is employed with a cross-entropy loss function. All the experiments are computed using a desktop computer system having 16 GB of RAM with an Nvidia GPU Card, and performed in the Pytorch \cite{paszke2019pytorch} environment, which is a popular machine learning library.

Table \ref{AF_CvRNN} gives the accuracy comparison of different activation functions for all the three data sets, where we have given the mean and standard deviation of accuracy for 10 different realizations. For MNIST, almost all activation functions perform well. It can be observed that the Swish, ReLU, Leaky ReLU, ELU, and Sine activation functions work well for the CIFAR-10 and CIFAR-100 data sets. Furthermore, in all three data sets, the adaptive and rowdy activation functions outperform. 

\section{Activation functions for physics-informed machine learning}
In recent years, physics-based machine learning (PIML) \cite{karniadakis2021physics} algorithms for scientific calculations have been revived at a considerably faster speed owing to the success of automatic differentiation (AD) \cite{baydin2018automatic}. These methods can solve not only the forward problems but also the notoriously difficult inverse problems governed by almost any type of differential equation. Various physics-informed machine learning algorithms have been proposed that can seamlessly incorporate governing physical laws into a machine learning framework. The SINDy framework \cite{brunton2016discovering} for analyzing dynamical systems, physics-informed machine learning for modeling turbulence \cite{wang2017physics}, physics-guided machine learning framework \cite{ karpatne2017theory}, the Neural ODE \cite{chen2018neural}, and most recently physics-informed neural networks (PINNs) \cite{raissi2019physics} are just a few examples of the many physics-informed machine learning algorithms that have been proposed and successfully used for solving problems from computational science and engineering fields.

Most of our previous discussion about activation function was centered around various classification problems in the computer science community. In this section, we focus on the activation functions for the PIML framework. Apart from the accurate and efficient AD algorithm, PIML methods also need tailored activation functions where the output of the network not only satisfies the given data but also satisfies the governing physical laws in the form of differential equations where various spatio-temporal derivative terms are involved. In this work, we are using PINNs for our investigation. PINN is a very efficient method for solving both forward and ill-posed inverse differential equations involving sparse and multi-fidelity data. The main feature of the PINN is that it can seamlessly incorporate all the information, like governing equation, experimental data, initial/boundary conditions,
etc., into the loss function, thereby recasting the original PDE problem into an optimization problem. PINNs achieved remarkable success in many applications in science and engineering; see \cite{raissi2019deep, jagtap2022physics, raissi2020hidden, shukla2021physics, jagtap2020extended, jagtap2020conservative, shukla2021parallel, jagtap2022deeplearning, hu2021extended}, and the references therein for more details. The first comprehensive theoretical analysis of PINNs for a prototypical nonlinear PDE, the Navier-Stokes equations, has been presented in \cite{de2022error}.

\subsection{Differentiability}
\begin{figure}[!h]
\centering
\includegraphics[scale=0.52]{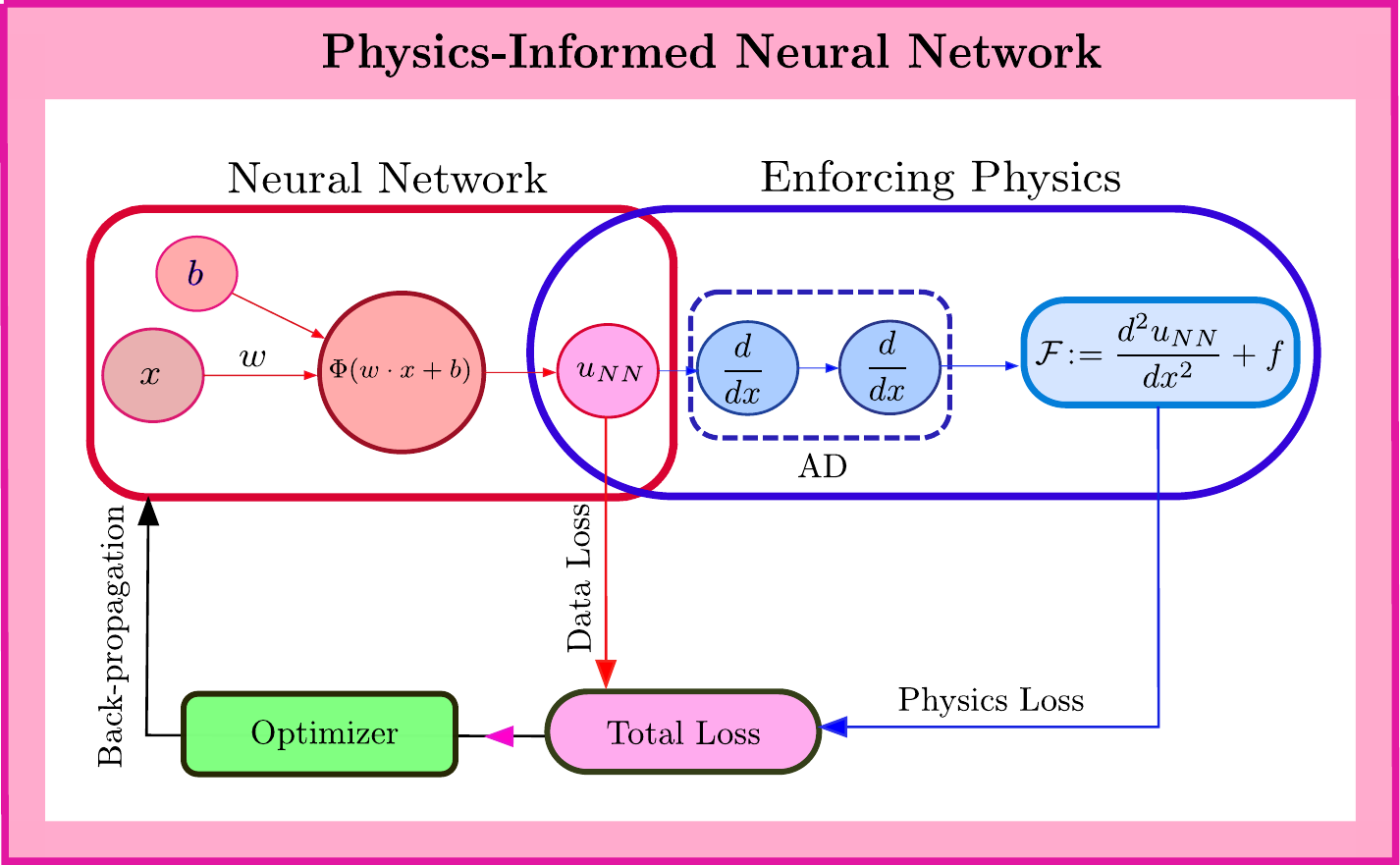}
\caption{Left part: One-hidden layer, single neuron, feed-forward neural network, and right part: physics-informed part where neural network's output is forced to satisfy the governing physical laws. In particular, the automatic differentiation (AD) is employed to find the derivatives present in the governing differential equation.}
\label{fig:AF_PINN}
\end{figure}

Differentiability is an important property of the activation function. The activation function must be differentiable everywhere.
Consider the example of one-dimensional Laplace equation with forcing $f$ as $$\frac{d^2}{dx^2}u = -f,  ~ x \subset \mathbb{R}.$$ Here we use a one-hidden-layer neural network with a single neuron, see figure \ref{fig:AF_PINN}. 
The network's input is the independent variable $x$, and its output is $u_{NN} \approx u(x)$. The network output $u_{NN}$ satisfies both data as well as physics, as shown in the figure. The loss function is given as
$$\mathcal{L} (\Theta) = \sum_{j=1}^{N_u} |u^j_{NN}(\Theta) - u_{\text{data}}|^2 + \sum_{j=1}^{N_f} |\mathcal{F}^j(\Theta)|^2, $$
where the first term on RHS is the data mismatch loss where $N_u$ are the number of data points, and the second term is the residual of the ODE given as $\mathcal{F} \triangleq \frac{d^2}{dx^2}(u_{NN}) -f$ with $N_f$ collocation points. The symbol $\Theta$ represents all the tunable parameters in the networks, such as weights, biases, etc.
Let $\Phi$ be the activation function that is applied before passing the output of the neuron. Then, the network output can be given as
$$ u_{NN} = \Phi(w\cdot x + b ).$$
The network output $u_{NN}$ not only satisfies the given data obtained from the boundary conditions or the experimental results, but also satisfies the physical laws (ODE residue), as shown by the right-hand side blue box where the AD is employed to differentiate the network's output.
It can be seen that the activation function plays an important role in order to correctly enforce the physical laws. In this case we required the value of $$\frac{d^2}{dx^2} (u_{NN}) =  \frac{d^2}{dx^2} \Phi(w\cdot x + b ).$$
This requirement imposes several restrictions on the activation functions, especially high-order of continuity. In particular, activation functions such as step function, ReLU, and leaky ReLU are not good candidates for solving such problems for the obvious reason of the non-existence of their derivatives in particular locations.

The most commonly used activation function that has been successfully employed in physics-informed machine learning frameworks is the hyperbolic tangent function. In recent years, sinusoidal activation functions have been shown to be beneficial for certain PDEs \cite{jagtap2020adaptive}.
In this section, we systematically compare the performances of different activation functions for the following three partial differential equations where the order of derivative increases from the first to the last case.

\vspace{0.5cm}
\noindent \textbf{Case 1 : Linear Convection Equation (LCE)} is a hyperbolic conservation laws given as
$$u_t + c u_x = 0,~ x\in[0,2], t>0,$$
where $c = 1.0$ is the wave speed. If the initial profile is $u (0, x) = u_0(x)$, then the exact solution is $u (t, x) = u_0(x-ct) $. The initial condition is a square pulse, which moves in space from left to right without changing its shape.

\vspace{0.5cm}
\noindent \textbf{Case 2 : Viscous Burgers Equation} is given as
$$u_t + u u_x = \nu u_{xx},~ x\in[-1,1], t>0,$$
where $c = 1.0$ is the wave speed. $u(0,x) = -\sin(\pi x)$ is the initial profile, and the exact solution is obtained using spectral methods. The viscosity coefficient $\nu = 0.01/\pi$. Due to the dominant nonlinear convection term, the viscous Burgers gives a high gradient solution.

\vspace{0.5cm}
\noindent \textbf{Case 3 : The Boussinesq Equation} with Dirichlet boundary and initial conditions is given by
$$
u_{tt} - u_{xx} - u_{xxxx} - 3(u^2)_{xx} = 0,~ x\in [-20, 20], t \in [-5, 5],
$$
with $u(x, -5) = u_0(x),$ is the initial condition. 
The exact solution is assumed to be of the form
$$u(x,t) = \frac{1}{2} \text{sech}^2 \left( \frac{x+t \sqrt{2}}{2}\right). $$

\begin{table*}[!h]
\centering
\begin{tabular}{cccccc} 
\hline
 &$\#$ Layers & $\#$ Neurons& $\#$ Residual Pts. & $\#$ Data Pts. &Learning rate \\ \hline
\textbf{Case 1} & 4& 30 & 7000& 150 &4e-04\\
\textbf{Case 2} & 6& 30 & 8000 & 200  &1e-04\\
\textbf{Case 3} & 6 &40 & 14000 & 300 &1e-04\\\hline
\end{tabular}
\caption{The hyperparameters used for three cases.}
\label{AF_PDE0}
\end{table*}

\begin{table*}[!h]
\centering
\begin{tabular}{cccc} 
\hline
 &\textbf{Case 1} & \textbf{Case 2}& \textbf{Case 3} \\ \hline
\textbf{Fixed Sine} & $\mathbf{3.545e-03 \pm 4.581e-04}$ & $5.894e-03 \pm 5.436e-04$ & $\mathbf{6.490e-03 \pm 3.846e-04}$\\
\textbf{Fixed GCU} & $4.553e-03 \pm 7.633e-04$ & $8.225e-03 \pm 3.736e-04$ & $8.095e-03 \pm 7.757e-04$\\
\textbf{Fixed Sigmoid} & $2.643e-02 \pm 7.536e-03$ & $1.967e-01 \pm 6.382e-02$ & $3.646e-01 \pm 6.547e-02$\\
\textbf{Fixed Tanh} & $4.989e-03 \pm 5.938e-04$ & $\mathbf{4.001e-03 \pm 4.504e-04}$& $6.546e-03 \pm 5.007e-04$\\
\textbf{Fixed Swish} & $6.224e-03 \pm 5.995e-04$ & $7.782e-03 \pm 6.648e-04$ & $7.648e-03 \pm 6.864e-04$\\
\textbf{Fixed ELU} ($\mathbf{\alpha = 1}$) & $1.549e-02 \pm 7.588e-03$  & - & -  \\\hline
\textbf{Adaptive Sine} & $2.255e-03 \pm 5.627e-04$ & $3.052e-03 \pm 4.368e-04$ & $\mathbf{1.370e-03 \pm 4.526e-04}$\\
\textbf{Adaptive GCU} & $3.908e-03 \pm 4.683e-04$ & $2.126e-03 \pm 3.877e-04$ & $1.875e-03 \pm 5.117e-04$\\
\textbf{Adaptive Sigmoid} & $1.052e-02 \pm 5.442e-03$ & $1.011e-01 \pm 4.637e-02$ & $1.992e-01 \pm 2.621e-02$\\
\textbf{Adaptive Tanh} & $\mathbf{1.927e-03 \pm 4.821e-04}$ & $\mathbf{1.731e-03 \pm 6.890e-04}$& $2.552e-03 \pm 6.909e-04$\\
\textbf{Adaptive Swish} & $4.652e-03 \pm 7.627e-04$ & $2.931e-03 \pm 7.526e-04$ & $3.526e-03 \pm 5.267e-04$\\
\textbf{Adaptive ELU} ($\mathbf{\alpha = 1}$) & $1.215e-02 \pm 5.007e-03$  & - & -  \\\hline
\textbf{Rowdy Sine} & $\mathbf{9.526e-04 \pm 8.226e-05}$ & $1.525e-03 \pm 3.526e-04$ & $\mathbf{8.356e-04 \pm 5.556e-04}$\\
\textbf{Rowdy GCU} & $1.262e-03 \pm 2.526e-04$ & $\mathbf{1.206e-03 \pm 4.524e-04}$ & $1.523e-03 \pm 9.442e-04$\\
\textbf{Rowdy Sigmoid} & $6.326e-03 \pm 9.767e-04$ & $6.257e-02 \pm 8.116e-03$ & $6.362e-02 \pm 5.562e-03$\\
\textbf{Rowdy Tanh} & $1.637e-03 \pm 3.960e-04$ & $1.902e-03 \pm 6.362e-04$& $1.049e-03 \pm 4.257e-04$\\
\textbf{Rowdy Swish} & $1.986e-03 \pm 6.398e-04$ & $2.526e-03 \pm 5.257e-04$ & $1.018e-03 \pm 3.466e-04$\\
\textbf{Rowdy ELU} ($\mathbf{\alpha = 1}$) & $3.499e-03 \pm 4.810e-03$  & - & -  \\\hline
\end{tabular}
\caption{Mean and standard deviation of relative $L_2$ errors for three cases using different fixed and adaptive activation functions. The \textit{Adaptive} and \textit{Rowdy} (with $K=5$) refers \cite{jagtap2020adaptive} and \cite{jagtap2022deep}, respectively. The column-wise best results are highlighted in bold for fixed, adaptive and Rowdy activation functions.}
\label{AF_PDE}
\end{table*}

\begin{figure}[!h]
\centering
\includegraphics[trim=0cm 1cm 0cm 0cm,  scale=0.55]{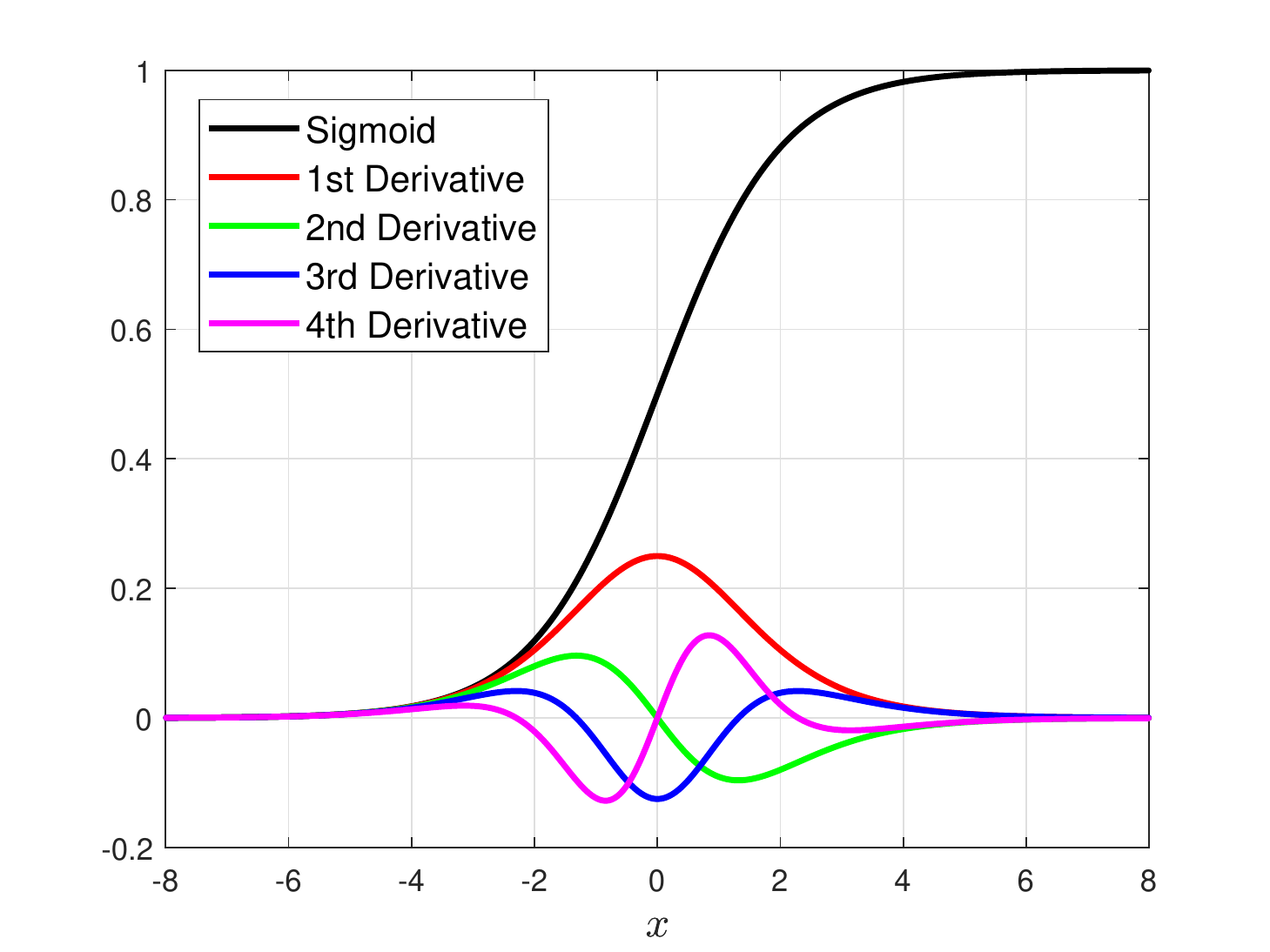}
\caption{The Sigmoid and its derivatives.}
\label{fig:SigDeri}
\end{figure}

Table \ref{AF_PDE0} shows all the hyperparameters that have been chosen for three different test cases. The number of data points represents both boundary as well as initial data. We used the first 10,000 iterations of the Adam \cite{kingma2014adam} optimizer followed by the L-BFGS \cite{byrd1995limited} optimizer till convergence. Table \ref{AF_PDE} displays the mean and standard deviation of the relative $L_2$ error for ten different realizations. It can be seen that sine, tanh, and Swish perform well in all cases, where as ELU's performance is poor in the first case. ELU can not be used for cases 2 and 3, as its higher order derivative does not exist. The sigmoid performs poorly, especially for PDEs involving high-order derivatives. This is obvious because higher order sigmoid derivatives have low amplitude, as shown in Figure \ref{fig:SigDeri}. Moreover, both adaptive and rowdy activations perform well compared to their fixed counterparts. The column-wise best performance of the activation function is shown using the bold symbols.

Although the fixed activation function gives better predictive accuracy, it is well known that the adaptive activation functions perform better than their fixed counterparts; see, for example, \cite{jagtap2020adaptive, jagtap2020locally}.

\subsection{Range}
The range of the activation function is another important aspect that needs further discussion. The field variable's range is occasionally known a priori when using the PIML framework to solve PDEs. For instance, the fluid's density will always be positive while simulating fluid flows (if there is no vacuum state present in the flow). The selection of an activation function for the output layer depends strongly on this information.
In the fluid example, activation can be used to produce only positive values by discarding all negative ones.
This effective selection of activation functions and incorporation of prior knowledge about the field variables has an immediate impact on the convergence of network training.  

\subsection{Performance comparison using different machine learning libraries}
Several machine learning libraries, such as TensorFlow \cite{abadi2016tensorflow}, PyTorch \cite{paszke2019pytorch}, and JAX \cite{frostig2018compiling} can be used to efficiently implement PINNs. Python is a more practical language to use for PINNs because it is the most popular programming language for machine learning, hence the majority of these packages are developed in Python.
Deep learning libraries such as PyTorch and Tensorflow offer high-level APIs for modern deep learning techniques. In contrast, the JAX library for arbitrary differentiable programming is more functionally oriented. JAX is a new high-performance machine learning library that combines  Autograd \cite{maclaurin2015autograd} and Accelerated Linear Algebra (XLA) \cite{frostig2018compiling}. The Autograd implementation in JAX enables automatic differentiation between native Python and NumPy routines. In addition to forward-mode differentiation, the Autograd engine in JAX also enables reverse-mode differentiation, and the two can be combined in any order. To get the benefits of static compilation and interpretation, just-in-time (JIT) compilation is used. For this, JAX also takes advantage of XLA to enhance the functionality of the compiled code.

In this work, we compare various activation functions for accurately evaluating the derivatives in the governing equations using TensorFlow (TF2), PyTorch, and JAX (with JIT) libraries.
\begin{table*}[!h]
\centering
\begin{tabular}{cccc} 
\hline
 &\textbf{TensorFlow} & \textbf{Pytorch}& \textbf{JAX} \\ \hline
\textbf{Fixed Sine} & $6.490e-03 \pm 3.846e-04$ & $7.086e-03 \pm 4.637e-04$ & $\mathbf{1.090e-03 \pm 2.860e-04}$\\
\textbf{Fixed GCU} & $5.125e-03 \pm 6.463e-04$ & $5.998e-03 \pm 3.657e-04$ & $\mathbf{1.346e-03 \pm 4.647e-04}$\\
\textbf{Fixed Sigmoid} & $3.646e-01 \pm 6.547e-02$ & $4.683e-01 \pm 5.483e-02$ & $\mathbf{8.638e-02 \pm 8.956e-03}$\\
\textbf{Fixed Tanh} & $6.546e-03 \pm 5.007e-04$ & $7.859e-03 \pm 8.750e-04$& $\mathbf{8.599e-04 \pm 9.907e-05}$\\
\textbf{Fixed Swish} & $7.648e-03 \pm 6.864e-04$ & $5.011e-03 \pm 7.486e-04$ & $\mathbf{9.053e-04 \pm 8.082e-05}$\\\hline
\textbf{Adaptive Sine} & $1.370e-03 \pm 4.526e-04$ & $3.372e-03 \pm 3.163e-04$ & $\mathbf{7.440e-04 \pm 6.060e-05}$\\
\textbf{Adaptive GCU} & $3.986e-03 \pm 5.647e-04$ & $2.575e-03 \pm 5.775e-04$ & $\mathbf{9.221e-04 \pm 8.836e-05}$\\
\textbf{Adaptive Sigmoid} & $1.992e-01 \pm 2.621e-02$ & $7.683e-02 \pm 9.083e-03$ & $\mathbf{4.638e-03 \pm 8.994e-03}$\\
\textbf{Adaptive Tanh} & $2.552e-03 \pm 6.909e-04$ & $2.157e-03 \pm 7.111e-04$& $\mathbf{5.489e-04 \pm 6.995e-05}$\\
\textbf{Adaptive Swish} & $3.526e-03 \pm 5.267e-04$ & $1.738e-03 \pm 4.738e-04$ & $\mathbf{6.053e-04 \pm 7.072e-05}$\\\hline
\textbf{Rowdy Sine} & $8.356e-04 \pm 5.556e-04$ & $1.098e-03 \pm 3.226e-04$ & $\mathbf{2.578e-04 \pm 2.860e-04}$\\
\textbf{Rowdy GCU} & $1.003e-03 \pm 1.858e-04$ & $2.012e-03 \pm 4.676e-04$ & $\mathbf{3.792e-04 \pm 3.094e-04}$\\
\textbf{Rowdy Sigmoid} & $6.362e-02 \pm 5.562e-03$ & $2.563e-02 \pm 5.256e-03$ & $\mathbf{3.678e-03 \pm 4.976e-04}$\\
\textbf{Rowdy Tanh} & $1.049e-03 \pm 4.257e-04$ & $8.861e-04 \pm 7.850e-05$& $\mathbf{3.798e-04 \pm 8.857e-05}$\\
\textbf{Rowdy Swish} & $1.018e-03 \pm 3.466e-04$ & $9.061e-04 \pm 8.659e-05$ & $\mathbf{2.748e-04 \pm 6.092e-05}$\\\hline
\end{tabular}
\caption{Clean Data: Performance comparison of the predictive accuracy of the solution of the Boussinesq equation using different ML libraries. We used different fixed and adaptive activation functions to compare the performance. The row-wise best results are highlighted in bold for all the activation functions.}
\label{AF_Imp}
\end{table*}
\begin{table*}[!h]
\centering
\begin{tabular}{cccc} 
\hline
 &\textbf{TensorFlow} & \textbf{Pytorch}& \textbf{JAX} \\ \hline
\textbf{Fixed Sine} & $1.037e-02 \pm 4.537e-03$ & $4.073e-02 \pm 7.411e-03$ & $\mathbf{7.570e-03 \pm 5.950e-04}$\\
\textbf{Fixed GCU} & $5.624e-02 \pm 7.378e-03$ & $2.457e-02 \pm 3.952e-03$ & $\mathbf{8.866e-03 \pm 4.273e-04}$\\
\textbf{Fixed Sigmoid} & $5.286e-01 \pm 6.077e-02$ & $5.983e-01 \pm 7.483e-02$ & $\mathbf{2.196e-01 \pm 2.956e-02}$\\
\textbf{Fixed Tanh} & $9.820e-03 \pm 1.607e-03$ & $3.314e-02 \pm 7.525e-03$& $\mathbf{7.021e-03 \pm 9.907e-04}$\\
\textbf{Fixed Swish} & $9.964e-03 \pm 2.726e-03$ & $8.521e-02 \pm 9.627e-03$ & $\mathbf{8.045e-03 \pm 8.082e-04}$\\\hline
\textbf{Adaptive Sine} & $9.362e-03 \pm 1.008e-03$ & $1.257e-02 \pm 7.004e-03$ & $\mathbf{4.526e-03 \pm 1.539e-03}$\\
\textbf{Adaptive GCU} & $8.053e-03 \pm 3.358e-04$ & $2.057e-02 \pm 1.641e-03$ & $\mathbf{5.647e-03 \pm 5.778e-04}$\\
\textbf{Adaptive Sigmoid} & $2.637e-01 \pm 5.821e-02$ & $2.376e-01 \pm 5.222e-02$ & $\mathbf{7.001e-02 \pm 4.526e-03}$\\
\textbf{Adaptive Tanh} & $8.003e-03 \pm 3.027e-03$ & $1.078e-02 \pm 5.373e-03$& $\mathbf{7.352e-03 \pm 7.087e-04}$\\
\textbf{Adaptive Swish} & $6.904e-03 \pm 8.637e-04$ & $3.637e-02 \pm 6.992e-03$ & $\mathbf{3.565e-03 \pm 9.227e-04}$\\\hline
\textbf{Rowdy Sine} & $6.226e-03 \pm 8.267e-04$ & $8.274e-03 \pm 1.860e-03$ & $\mathbf{2.523e-03 \pm 1.436e-03}$\\
\textbf{Rowdy GCU} & $7.644e-03 \pm 6.001e-04$ & $5.436e-03 \pm 9.570e-04$ & $\mathbf{3.444e-03 \pm 8.653e-04}$\\
\textbf{Rowdy Sigmoid} & $8.286e-02 \pm 6.137e-03$ & $8.364e-02 \pm 8.003e-03$ & $\mathbf{3.842e-02 \pm 2.956e-02}$\\
\textbf{Rowdy Tanh} & $2.536e-03 \pm 8.847e-04$ & $8.000e-03 \pm 6.141e-04$& $\mathbf{9.091e-04 \pm 1.737e-04}$\\
\textbf{Rowdy Swish} & $3.663e-03 \pm 6.756e-04$ & $9.032e-03 \pm 2.367e-03$ & $\mathbf{8.255e-04 \pm 3.530e-04}$\\\hline
\end{tabular}
\caption{Noisy data (5\% random Gaussian noise): Performance comparison of the predictive accuracy of the solution of the Boussinesq equation using different ML libraries. We used different fixed and adaptive activation functions to compare the performance. The row-wise best results are highlighted in bold for all the activation functions.}
\label{AF_Imp2}
\end{table*}
For this reason, we consider the Boussinesq equation because of the presence of a fourth-order derivative term. Moreover, we used the same hyperparameters defined earlier to solve the Boussinesq equation. The Adam optimizer is used for the first 10,000 iterations, followed by the L-BFGS optimizer till convergence. The mean and standard deviation of relative $L_2$ error for ten different realizations are shown in Table \ref{AF_Imp}, which compares the predictive accuracy of the solution using TensorFlow, PyTorch, and JAX libraries for clean data. It can be seen that compared to TensorFlow and Pytorch, JAX improves the predictive accuracy in all cases, including fixed, adaptive, and Rowdy activation functions.
We also compare the predictive accuracy of the solution for the noisy data. In particular, we have added 5 \% random Gaussian noise to the initial and boundary data sets to achieve this. Table \ref{AF_Imp2} shows this comparison, where it can be seen that the predictive accuracy degrades by almost an order of magnitude in all cases. Again, in all cases, compared to the Tensorflow and Pytorch libraries, JAX offers a substantial increase in accuracy (about an order of magnitude) in the solution of PDEs. 
\begin{figure*}[!h]
\centering
\includegraphics[trim=2cm 2cm 2cm 2cm,  scale=0.4]{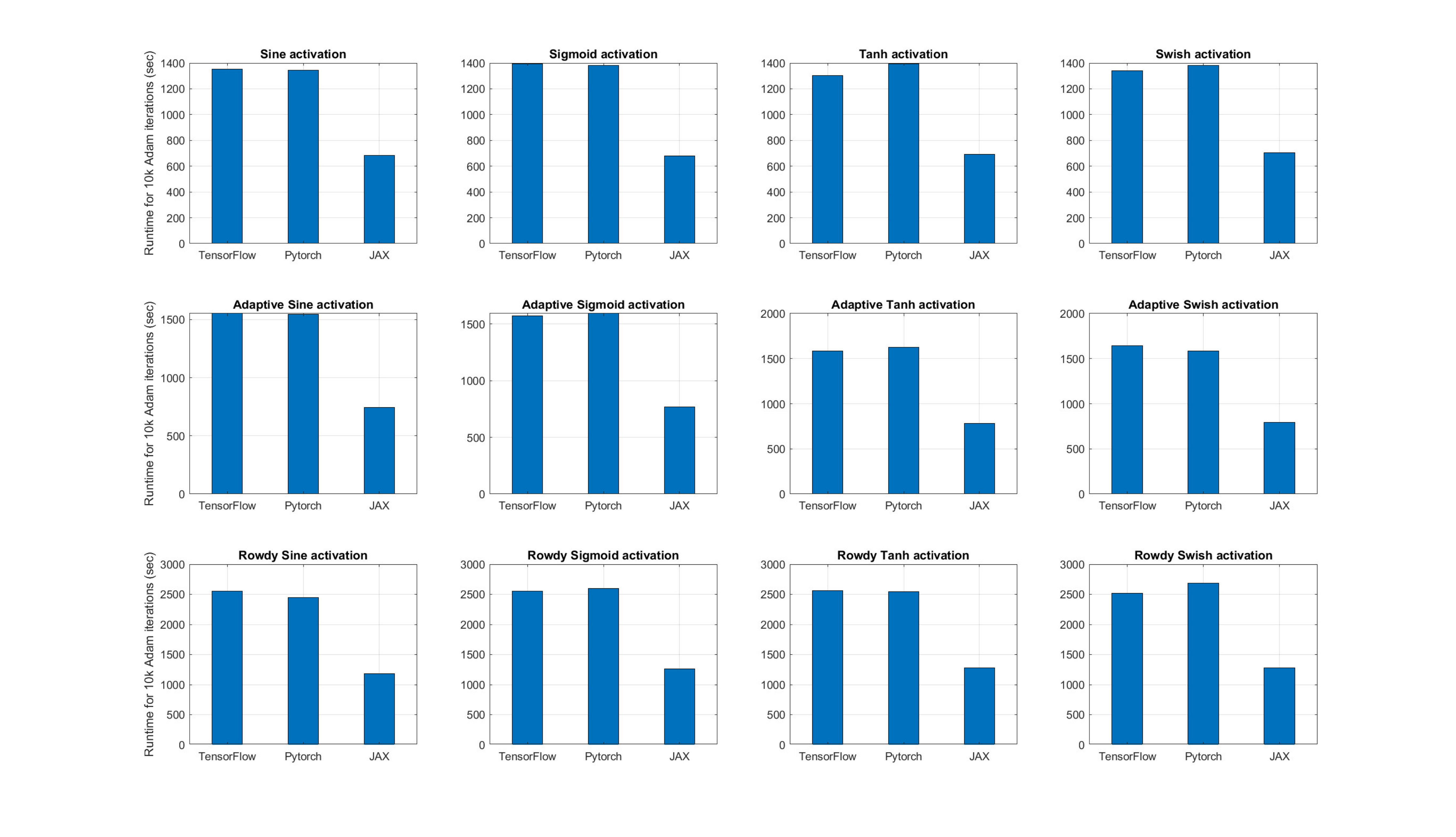}
\caption{TensorFlow, Pytorch, and JAX run-time evaluations for the first 10,000 Adam iterations of the Boussinesq equation with clean data. The top row consists of fixed activations, the middle row consists of adaptive activations, and the bottom row consists of Rowdy activations with $(K=2)$.}
\label{fig:RunTime}
\end{figure*}
The runtime comparison of TensorFlow, Pytorch, and JAX for the first 10,000 Adam iterations of the Boussinesq equation with clean data is shown in Figure \ref{fig:RunTime}. The top row consists of fixed activations, the middle row consists of adaptive activations, and the bottom row consists of Rowdy activations with $(K=2)$. It can be observed that adaptive activation increases the overall cost by 8-10\% compared to fixed activations. In the case of Rowdy activations, the cost increment is around 75-80\%. One remedy to overcome the cost associated with Rowdy activations is to use the transfer learning strategy discussed \cite{jagtap2022deep}. In all cases, JAX (with JIT) is computationally much more efficient than the TensorFlow (TF2) and Pytorch libraries.

Next, we compare different activation functions for function approximation using only data points.
\begin{table*}[htpb]
\centering
\begin{tabular}{ccc} 
\hline
  & \textbf{Function Approximation}& \textbf{Helmholtz Equation (JAX)} \\\hline
\textbf{Sine}  & $1.865e-1 \pm 4.463e-2$   & $\mathbf{1.032e-1 \pm 5.645e-2}$\\
\textbf{GCU}  & $2.106e-1 \pm 5.052e-2$   & $1.564e-1 \pm 4.995e-2$\\
\textbf{Sigmoid}  & $3.402e-1 \pm 4.052e-2$   & $4.958e-1 \pm 7.845e-2$\\
\textbf{Tanh}  & $1.859e-1 \pm 7.982e-2$   & $1.378e-1 \pm 4.284e-2$\\
\textbf{Swish}  & $\mathbf{8.327e-2 \pm 7.432e-2}$   & $2.734e-1 \pm 2.115e-2$\\
\textbf{ReLU}   & $1.742e-1 \pm 8.055e-2$   & -\\
\textbf{Leaky ReLU}   & $8.963e-1 \pm 9.012e-2$   & -\\
\textbf{ELU}   & $3.737e-1 \pm 9.012e-2$   & -\\\hline
\textbf{Adaptive Sine}  & $6.673e-2 \pm 5.715e-3$   & $8.553e-2 \pm 7.473e-3$\\
\textbf{Adaptive GCU}  & $4.709e-2 \pm 8.055e-3$   & $9.042e-2 \pm 6.647e-3$\\
\textbf{Adaptive Sigmoid}  & $5.245e-2 \pm 4.035e-3$   & $9.537e-2 \pm 4.365e-3$\\
\textbf{Adaptive Tanh}  &  $\mathbf{3.637e-2 \pm 7.982e-3}$   & $\mathbf{8.257e-2 \pm 5.737e-3}$\\
\textbf{Adaptive Swish}  &$ 4.626e-2 \pm 8.463e-3 $  & $9.952e-2 \pm 9.115e-3$\\
\textbf{Adaptive ReLU}   & $8.436e-2 \pm 1.687e-3$   & -\\
\textbf{Adaptive Leaky ReLU}   & $6.362e-2 \pm 4.239e-3$   & -\\
\textbf{Adaptive ELU}   & $7.362e-2 \pm 8.603e-3$   & -\\\hline
\textbf{Rowdy Sine}  & $6.478e-3 \pm 8.858e-4$   & $\mathbf{4.872e-2 \pm 5.486e-3}$\\
\textbf{Rowdy GCU}  & $7.997e-3 \pm 5.857e-4$   & $7.473e-2 \pm 4.778e-3$\\
\textbf{Rowdy Sigmoid}  & $5.245e-3 \pm 4.035e-4$   & $6.462e-2 \pm 2.084e-3$\\
\textbf{Rowdy Tanh}  & $4.971e-3 \pm 3.745e-4$   & $5.137e-2 \pm 5.677e-3$\\
\textbf{Rowdy Swish}  &$\mathbf{3.245e-3 \pm 4.035e-4}$   & $6.932e-2 \pm 6.354e-3$\\
\textbf{Rowdy ReLU}   & $7.985e-3 \pm 1.031e-3$  &-\\
\textbf{Rowdy Leaky ReLU}   & $5.342e-3 \pm 8.259e-4$   & -\\
\textbf{Rowdy ELU}   & $6.362e-3 \pm 9.534e-4 $  & -\\
\hline
\end{tabular}
\caption{Comparison of different activation functions for function approximation and PINNs. The \textit{Adaptive} and \textit{Rowdy} (with $K=5$) refers \cite{jagtap2020adaptive} and \cite{jagtap2022deep}, respectively. The column-wise best results are highlighted in bold for fixed, adaptive, and Rowdy
activation functions.}
\label{AF_CvR2}
\end{table*}
The following discontinuous function is
approximated by a feed-forward neural network.
$$f(x) =\begin{cases}
0.2 ~\sin(6x) & x \leq 0, \\ 1 + 0.1~ x \cos(22x) & \text{Otherwise}.
\end{cases}$$
This function contains both high and low frequency components along with the discontinuity at $x=0$. The domain is $[-3, 3]$ and the number of randomly chosen training points used is 300. We also solved the Helmholtz equation in two dimensions, given by
$$ u_{xx} + u_{yy} + k^2 u = f(x,y),$$
where the following high-frequency solution is assumed $u(x,y) = \sin(8\pi x)\, \sin(16\pi y)$, with $k=1$. Table \ref{AF_CvR2} gives the relative $L_2$ error using different activation functions for the  function approximation with 10,000 iterations of Adam, and the Helmholtz equation with 40,000 Adam iterations. In particular, we used fixed activation, adaptive activation \cite{jagtap2020adaptive}, and Rowdy activation \cite{jagtap2022deep} with $K = 5$. The Rowdy activation performance is much better than the rest of them.

\section{Summary}
The most crucial element in a neural network that determines whether or not a neuron will be engaged and moved to the next layer is the activation function.
Simply said, it will determine whether or not the neuron's input to the network is significant to the prediction process.
The activation function plays a vital role in the learning process of a neural network. The ideal activation function, which would fit every model and produce accurate results, does not exist. Thus, the selection of the proper activation function is crucial for better accuracy and faster convergence of the network. In the literature, various activation functions are discussed, and most of them are inspired by the behavior of biological neural networks. To the best of our knowledge, this is the first thorough examination of activation functions for classification and regression problems. In addition, we presented a number of original contributions, which are discussed below.

\begin{itemize}
\item We discussed various classical (fixed) activation functions that have been commonly used, such as sigmoid, hyperbolic tangent, ReLU and their variants, and so on. In particular, we discussed the various merits and limitations of these activation functions. It is well known that the ReLU and its variants achieve state-of-the-art performance in many image classification problems. 
\item We also proposed a taxonomy of activation functions based on their applications, which is more suitable from a scientific computation point-of-view.  In particular, we discussed various complex-valued activation functions that have been employed in many important scientific fields such as electro-magnetics, bioinformatics, acoustics, etc. The discussion of quantized activations for both real and complex-valued activations is given briefly. The quantized activations are much more efficient than their full-precision activation counterparts, with a similar level of accuracy. With such quantized activation, it is possible to tackle big data and models much more easily and could be used in edge computing \cite{shi2016edge}.
\item The adaptive nature of the activation function can be achieved in several ways. Although these adaptive activations are more expensive, they can achieve better performance than their fixed counterparts. The state-of-the-art adaptive activation functions that can adapt as per the requirements are thoroughly discussed. There are several ways of adapting the activation functions, such as introducing tunable parameters or employing ensemble techniques, in which an ensemble of more than one activation function produces better results. Adaptive activation functions can also be developed by introducing stochastic parameters as well as fractional derivatives. Adaptive activation outperforms its classical counterparts for almost any problem. Furthermore, we also compared the performance of various fixed and adaptive activations for different classification data sets such as MNIST, CIFAR-10, and CIFAR -100 using MobileNet \cite{howard2017mobilenets} and VGG16 \cite{simonyan2014very} architectures, which are CNN based models. For the MNIST data set, almost all activation functions perform well. Also, it can be observed that activation functions such as swish, ReLU, Leaky ReLU, ELU, and sine perform well for CIFAR-10 and more difficult CIFAR-100 test cases. Also, the adaptive and Rowdy activation functions perform well for all the data sets.
\item In recent years, a physics-informed machine learning framework has emerged as an effective approach to scientific machine learning. Such methods have proved to be very successful for ill-posed inverse problems, which are difficult to solve using traditional methods. To this end, we discussed various requirements for activation functions that have been used in the physics-informed machine learning framework.
Accurate evaluation of spatio-temporal derivatives is important to accurately enforcing the physical laws. We compared the accuracy of different activation functions using various machine learning libraries such as TensorFlow, Pytorch, and JAX. In all cases, the adaptive and Rowdy activation functions perform well with both clean and noisy (with Gaussian noise) data sets. When solving PDEs, JAX offers a significant improvement in the predictive accuracy of solution for all activation functions (by an order of magnitude) over the Tensorflow and Pytorch libraries. Moreover, JAX (with a JIT compiler) is computationally very efficient compared to the Tensorflow and Pytorch libraries. Another advantage of JAX is that higher-order optimization strategies like AdaHessian \cite{yao2021adahessian} are significantly more easily implemented thanks to Accelerated Linear Algebra’s (XLA’s) ability to compute Hessians noticeably faster than other ML libraries.

Multi-scale modeling is an important area where PIML-based solvers can be proved effective. Multi-scale modeling is a type of modeling where several models at various scales (different frequencies) are used to represent a system simultaneously. The various models often concentrate on various resolution scales. Sometimes they derive from distinct kinds of physical rules. The necessity for multi-scale modeling typically results from the inadequacy of the available macroscale models and the inadequacy or excess information provided by the microscale models. One aims to achieve a reasonable compromise between accuracy and efficiency by merging the two points of view.
As discussed in \cite{jagtap2022deep}, the fixed activation function does not perform well when the solution contains high frequencies. Such a problem is called the \textit{spectral bias} problem. The \textit{Rowdy} activation functions have the ability to overcome the spectral bias problem by injecting sinusoidal noise over the base activation functions. 

Another interesting observation is that the activation functions such as ReLU and its other variants, including ELU, which perform well for classification tasks like MNIST, CIFAR-10 and CIFAR-100 , do not perform well for PIML-based regression problems due to the non-existence of their derivatives. The activation functions such as sine, hyperbolic tangents, and swish perform well for the regression problems. 
\end{itemize}

\subsection{Some future directions}
Recently, there has been a surge in the usage of machine learning for solving many scientific and engineering problems, and it is well-known that the model size increases exponentially with the complexity of the problems. Additionally, the data that may be made available through a range of sources does not suffer from scarcity. Thus, one of the major challenges is to handle such big models and data involving billions of parameters. Faster convergence of the networks' training can be achieved by using efficient and optimized activation functions, such as adaptive activations. Although more expensive, adaptive activation offers higher performance, and its effectiveness for both classification and regression tasks has been demonstrated across all machine learning domains. One way to mitigate the cost associated with adaptive activations is to use \textit{quantized adaptive activation functions}. The quantized nature of adaptive activation will certainly prove effective for big models involving large data sets.
Another direction is to further improve the performance of adaptive activation functions by applying new ideas to adapt these activations. One approach is to include '\textit{influential}' adaptive parameters that change the topology of the loss landscape dynamically and thus avoid local minima. These influential parameters are the ones that can generate multiple hypersurfaces as needed, effectively removing the saturation region. Spiking neural networks (SNN), which predominantly employ spiking neurons of the Leaky-Integrate-and-Fire (LIF) type, represent yet another area of research. Although ANN is the most effective artificial intelligence model in practical applications, SNNs are the most biologically plausible technology.
The mapping between the parameters of the LIF and the activation function employed in ANNs would be challenging to understand \cite{lu2022linear}. This would be useful to develop more efficient and universal activation functions.

Although there are many better adaptive activation functions present in the literature, the quest for finding the best activation function is, and will continue to be, an active area of research.

\section*{Acknowlegments}
This work was supported by the OSD/AFOSR Multidisciplinary University Research Initiative (MURI) program grant number FA9550-20-1-0358, and U.S. Department of Defence's Vannevar Bush Faculty Fellowship award number N00014-22-1-2795.

\bibliographystyle{unsrt}
\bibliography{ref}

\end{document}